\documentclass[fleqn,10pt]{wlscirep}

\usepackage{spverbatim}
\usepackage[table]{xcolor} 
\usepackage{booktabs}
\usepackage{hyperref}
\usepackage[utf8]{inputenc}
\usepackage[T1]{fontenc}
\usepackage{lineno}
\usepackage{float}

\usepackage{svg}
\usepackage{enumitem}
\usepackage{appendix}  
\usepackage{bbm}
\usepackage{wrapfig}

\usepackage{csquotes}

\setlist{nosep}

\usepackage{multirow}  

\usepackage{graphicx}
\usepackage{subcaption}

\usepackage{tikz}  

\usepackage{listings}

\usepackage{tabularx}

\newfloat{codebox}{htbp}{loc}[section]
\floatname{codebox}{Listing}

\colorlet{punct}{red!60!black}
\definecolor{background}{HTML}{EEEEEE}
\definecolor{delim}{RGB}{20,105,176}
\colorlet{numb}{magenta!60!black}

\lstdefinelanguage{json}{
    basicstyle=\normalfont\ttfamily\small,
    numbers=left,
    numberstyle=\scriptsize,
    stepnumber=1,
    numbersep=8pt,
    showstringspaces=false,
    breaklines=true,
    frame=lines,
    backgroundcolor=\color{background},
    literate=
     *{0}{{{\color{numb}0}}}{1}
      {1}{{{\color{numb}1}}}{1}
      {2}{{{\color{numb}2}}}{1}
      {3}{{{\color{numb}3}}}{1}
      {4}{{{\color{numb}4}}}{1}
      {5}{{{\color{numb}5}}}{1}
      {6}{{{\color{numb}6}}}{1}
      {7}{{{\color{numb}7}}}{1}
      {8}{{{\color{numb}8}}}{1}
      {9}{{{\color{numb}9}}}{1}
      {:}{{{\color{punct}{:}}}}{1}
      {,}{{{\color{punct}{,}}}}{1}
      {\{}{{{\color{delim}{\{}}}}{1}
      {\}}{{{\color{delim}{\}}}}}{1}
      {[}{{{\color{delim}{[}}}}{1}
      {]}{{{\color{delim}{]}}}}{1},
}

\lstdefinestyle{jsonstyle}{
  basicstyle=\ttfamily\small,
  breaklines=true,
  frame=single,
  showstringspaces=false,
  stringstyle=\color{blue!60!black},
  commentstyle=\color{gray},
  keywordstyle=\color{black},
}

\usepackage{colortbl}

\usepackage{longtable}

\RequirePackage{silence}
\usepackage{setspace}
\usetikzlibrary{arrows.meta, positioning, shapes.geometric, fit, calc}
\usepackage{adjustbox}

\definecolor{ibm0}{RGB}{100, 143, 255}   
\definecolor{ibm1}{RGB}{120, 94, 240}    
\definecolor{ibm2}{RGB}{220, 38, 127}    
\definecolor{ibm3}{RGB}{254, 97, 0}      
\definecolor{ibm4}{RGB}{255, 176, 0}     
\definecolor{color_human}{HTML}{648FFF}
\definecolor{color_llm}{HTML}{FE6100}
\tikzset{
    base/.style={draw, text centered, minimum height=3.8em, line width=0.7pt},
    startnode/.style={base, rectangle, rounded corners, fill=ibm0!25, draw=ibm0, text width=3.6cm, font=\large\bfseries},
    endnode/.style={base, rectangle, rounded corners, fill=ibm2!25, draw=ibm2, text width=3.6cm, font=\large\bfseries},
    llmcall/.style={base, rectangle, fill=ibm1!15, draw=ibm1, text width=5.2cm, align=center, font=\large},
    modellm/.style={base, rectangle, fill=color_llm!35, draw=color_llm, text width=5.2cm, align=center, font=\large},
    modehuman/.style={base, rectangle, fill=color_human!35, draw=color_human, text width=5.2cm, align=center, font=\large},
    session/.style={base, rectangle, fill=ibm0!15, draw=ibm0, text width=4.4cm, align=center, font=\large},
    decisionnode/.style={base, diamond, aspect=2, fill=ibm4!40, draw=ibm4!80!black, font=\normalsize, inner sep=3pt},
    choice/.style={base, rectangle, fill=ibm3!20, draw=ibm3, text width=3.6cm, font=\normalsize, align=center},
    datanode/.style={base, trapezium, trapezium left angle=70, trapezium right angle=110, fill=ibm0!10, draw=ibm0, text width=3.2cm, align=center, font=\large},
    savenode/.style={base, rectangle, fill=ibm1!20, draw=ibm1, text width=4.8cm, align=center, font=\large\bfseries},
    jsonbox/.style={rectangle, draw=gray!50, dashed, fill=gray!8, text width=4.4cm, font=\small, minimum height=1.6em, align=center},
    connector/.style={draw, -{Stealth[scale=1]}, line width=0.25pt},
    sideconn/.style={draw, dashed, -Latex, color=gray!60, line width=0.5pt}
}

\restylefloat{table}

\title{Math Education Digital Shadows for Investigating Learning with GenAI: Mathematics Performance, Anxiety, and Confidence in LLMs}
\author[1]{Naomi Esposito}
\author[1]{Anthony Tricarico}
\author[1]{Luisa Porzio}
\author[1,*]{Ali Aghazadeh Ardebili}
\author[1]{Massimo Stella}
\affil[1]{CogNosco Lab, University of Trento, Department of Psychology and Cognitive Science, Trento, Italy}
\affil[*]{Corresponding author: a.a.ardebili@unitn.it}

\begin{abstract}

Understanding the impact of large language models (LLMs) on mathematics education requires data on LLMs' mathematical performance and biases. To this end, we introduce Math Education Digital Shadows (MEDS), a dataset mapping how LLMs reason about mathematics across human- and AI-like personifications. MEDS comprises 28,000 runs from 14 LLMs (i.e., Mistral, Qwen, DeepSeek, IBM Granite, Microsoft Phi, and xAI Grok) generated under human-shadow and AI-assistant conditions. Each record (digital shadow) includes a set of prompts; psychological and sociodemographic metadata; and four mathematics tasks: (i) interviews about relationships with mathematics, (ii) three psychometric questionnaires on mathematics self-efficacy and anxiety, (iii) one cognitive network capturing attitudes towards mathematics, and (iv) 18 high-school mathematics quiz items enriched with reasoning explanations and confidence scores. Analyses of the data show that LLMs exhibit differences in attitudes and performance across human-shadow and AI-assistant modes, including human-like negative attitudes towards mathematics, logical fallacies, and overconfidence in mathematics. As a data resource, MEDS can benefit learning scientists and developers of safer AI tutors in mathematics.

\end{abstract}
\begin{document}

\flushbottom
\maketitle

\thispagestyle{empty}

Generative AI (GenAI) is making its way into mathematics education during a period of genuine pedagogical challenge \cite{giannakos2025promise,kasneci2023chatgpt}. The advent of large language models (LLMs) has given students around-the-clock access to expert-level systems \cite{gabriel2025pragmatic}, potentially broadening access to personalised educational content \cite{yan2024promises,bastani2025generative} while also heightening the risk of exposure to biased or fallacious educational material \cite{kasneci2023chatgpt,bender2021dangers,DeDuro2025,qiu2026information} or undermining students' self-efficacy \cite{bastani2025generative}. Recent LLM-auditing work on DeepSeek \cite{qiu2026information} shows why prompt-conditioned datasets should be used to compare model outputs across controlled conditions, especially when final answers may differ from intermediate reasoning or suppress specific information. This risk must be considered alongside GenAI's usage statistics. In the 2025 EU-wide Eurostat module on GenAI, including ChatGPT, Copilot, Gemini, and LLaMA, $9.4\%$ of people aged 16--74 reported using such tools for formal education \cite{eurostat2026euai,eurostat2025questionnaire}. Among individuals aged 16--24, educational usage rose to $39.3\%$, indicating that AI-assisted education is already a substantial practice among younger learners in Europe \cite{eurostat2026youthai}. This rapid adoption is also due to the speed at which LLMs are evolving \cite{kasneci2023chatgpt,wenger2026large}. Originally conceived as models for generating \cite{bender2021dangers} or summarising language \cite{zhang2024benchmarking}, LLMs are increasingly becoming agentic \cite{Chen2026,chen2026openclaw,qiu2026information}, i.e., capable of making decisions or giving feedback under uncertainty.

The evolution of LLMs towards agentic capabilities has also included significant improvements in mathematical problem solving. Whereas older models such as GPT-3 achieved very low scores ($14\%$) on high-school-level mathematics items (cf. MathBench \cite{liu2024mathbench}), more recent models such as GPT-4o achieved scores almost four times as high. Similar improvements in mathematical reasoning were also observed on other mathematics benchmarks \cite{balunovic2025matharena,benedetto2024using,gupta2025beyond}. Consequently, mathematics education is benefiting considerably from LLMs' improved feedback skills \cite{gabriel2025pragmatic}. Recent human-centred studies indicate that mathematics students rely on LLMs to elucidate problem-solving algorithms \cite{stohr2024perceptions}, evaluate solutions \cite{wang2025influence,yan2024promises}, and minimise the effort required to tackle challenging problems \cite{gabriel2025pragmatic,bastani2025generative}. Conversely, institutions are slowly embracing LLMs as mathematics tutors, e.g., GenAI assistants primarily devoted to supporting students' mathematics learning and problem solving \cite{giannakos2025promise,akheel2025guardrails}. This slow adoption rate is also due to a lack of clear evidence about the cognitive repercussions of using LLMs in mathematics education \cite{gupta2025beyond,wang2025influence}. This is a significant gap because mathematics entails not only linguistic and mathematical knowledge but also several human skills, such as confidence \cite{gabriel2025pragmatic}, anxiety management \cite{hopko2003abbreviated}, self-efficacy \cite{nielsen2003psychometric}, and resilience to failure \cite{may2009mathematics}. Therefore, allowing students to interact with potential LLM-based mathematics tutors raises a key research question (RQ):

\begin{center}
\textit{RQ: Can LLMs convey mathematical knowledge effectively without expressing mathematics anxiety or conveying disruptive mathematical biases to users, even across a wide variety of prompting conditions?}
\end{center}

To address this RQ, we introduce the Math Education Digital Shadows (MEDS) dataset as a reusable, multimodal, LLM-based resource. Exploring LLMs' mathematics performance and attitudes requires a representation that preserves not only problem-solving answers but also the contextual framings under which those answers appear, e.g., whether an LLM introduces stereotypical perceptions such as ``math is boring'' in its responses \cite{abramski2023cognitive}. Digital shadowing \cite{Bergs2021} offers a convenient way to achieve that representation. While digital twins are characterised by a bidirectional data flow with their physical counterparts, digital shadows are more parsimonious \cite{Ardebili2021EIOT,aghazadeh2024digital}: digital shadows record what a system does across controlled conditions without updating, simulating, or influencing external counterparts \cite{GAFFINET2025104230,Singh2021,ardebili2021digitaltwins}. To address this gap, we present MEDS as a digital-shadow resource that facilitates the study of how LLMs engage with mathematics under psychologically and educationally grounded conditions.

MEDS goes beyond mathematical performance by adding data on LLMs' confidence, anxiety, semantic associations, and reflective discourse about mathematics, AI tutoring, and STEM. We define a Math Education Digital Shadow as a structured record of an LLM's mathematics-related psychological profile and problem-solving behaviour generated under specified constraints (four tasks, prompting mode, and persona profile). Throughout this resource, we use the term \textbf{human shadow} to denote an LLM instance prompted to simulate a synthetic individual defined by sociodemographic and psychological attributes (as opposed to an AI assistant, in which the model responds as itself). The term persona is reserved strictly for the underlying JSON attribute dictionary that defines a human shadow, not for the simulated entity itself. The accompanying dataset documentation describes personas through sociodemographic features \cite{carrillo2026talk2ai} (e.g., age, gender, sexual orientation, city of residence, employment status, education level, and parental education) and psychological attributes (e.g., Big Five personality traits \cite{john1999big}, favourite subjects \cite{ciringione2025math}, and psychometric scores \cite{russell2026ultimate}). This structure makes latent human variables explicit and experimentally manipulable within LLMs. MEDS includes four complementary tasks used in relevant previous STEM-education studies: (i) qualitative responses about mathematics and mathematical thinking, inspired by previous research on LLMs' affective biases \cite{de2025measuring}; (ii) psychometric ratings with textual justifications \cite{may2009mathematics,nielsen2003psychometric,hopko2003abbreviated}; (iii) semantic associations with valence labels \cite{stella2019forma,abramski2023cognitive}; and (iv) mathematical problem solving with reasoning and confidence scores \cite{kranzler1997exploratory}.

\begin{figure}[h!]
    \centering
    \includegraphics[width=\linewidth]{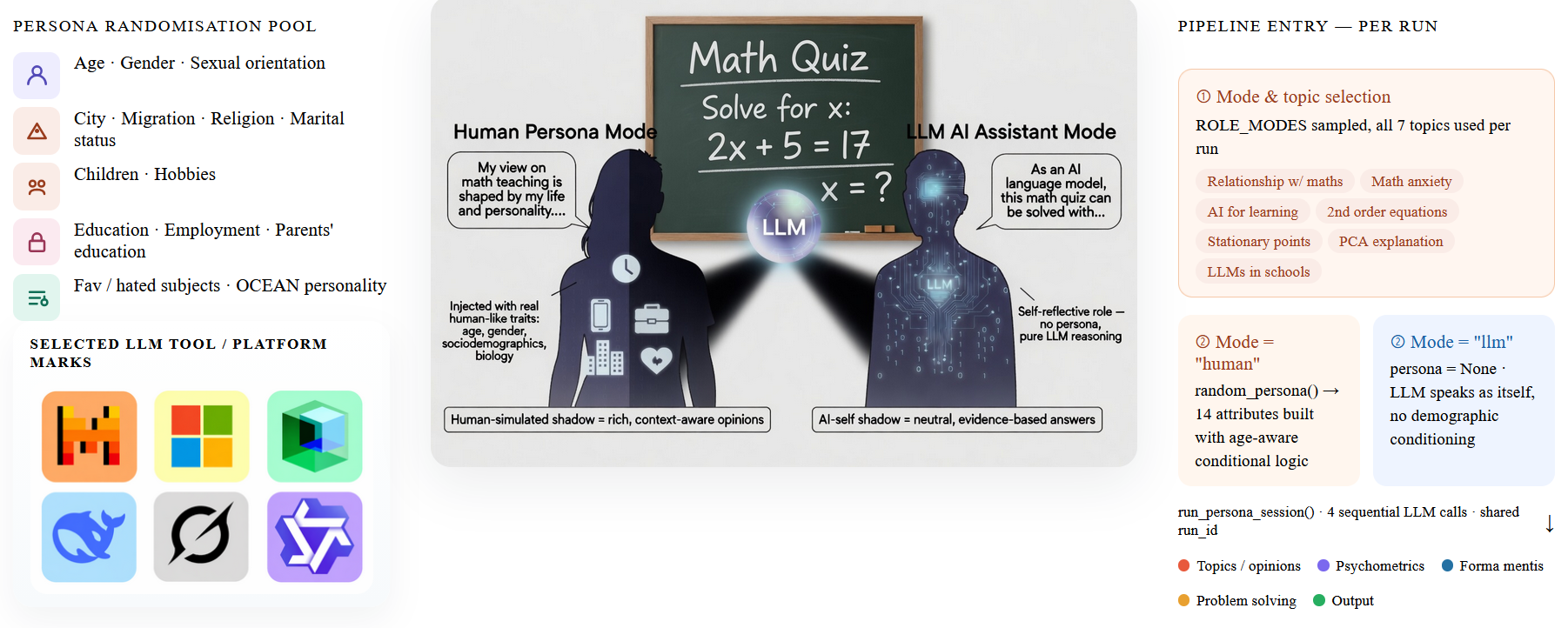}
    \includegraphics[width=\linewidth]{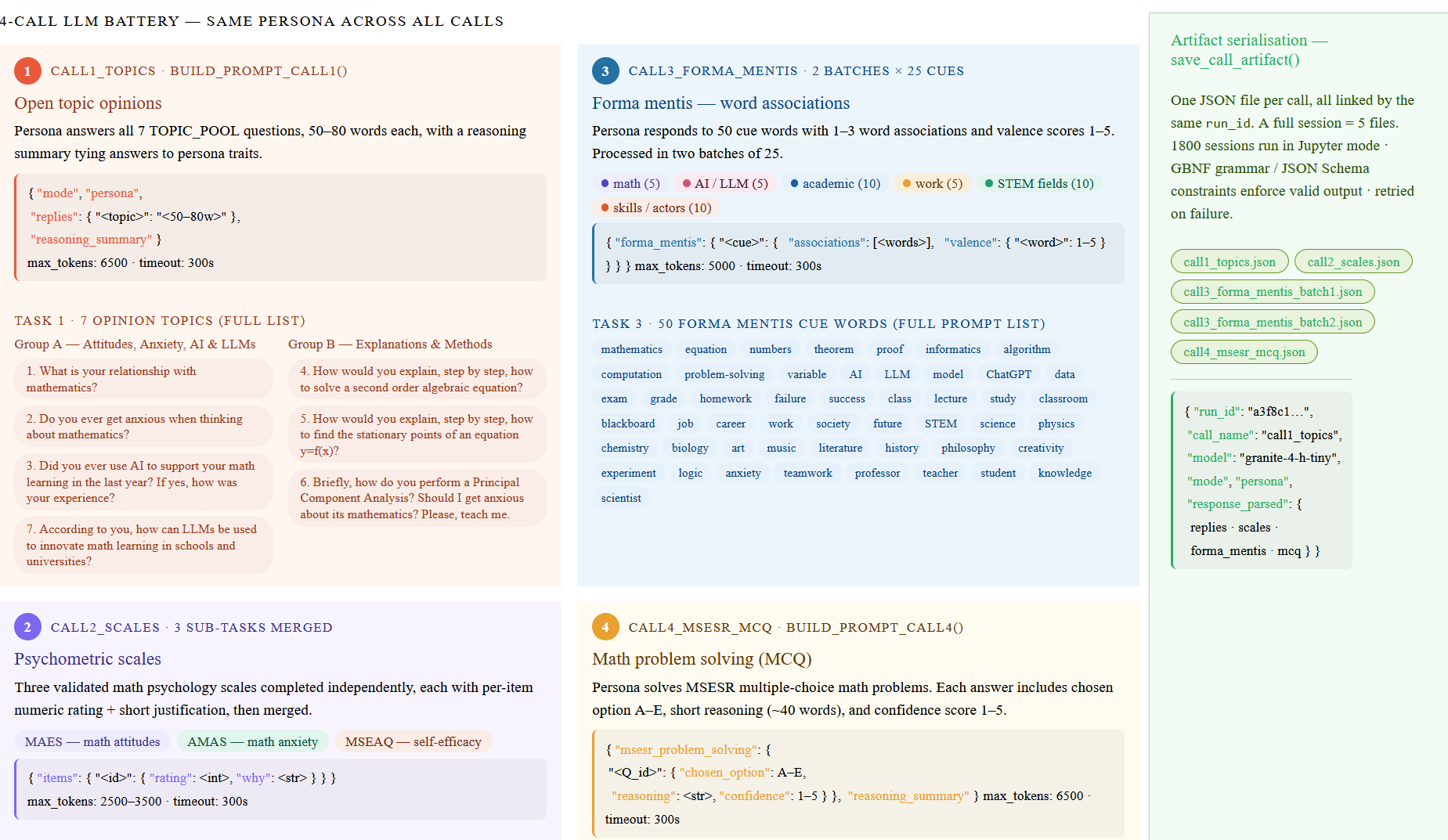}
    \caption{Overview of the MEDS prompting framework and response structure, showing the two modes (human shadows and AI assistants) and the four sequential tasks, together with example questions, scale items, and cue words used to generate each digital shadow. The label ``MAES'' in call 2 refers to the MSES scale. The \texttt{mode} field indicates human shadows (\texttt{"human"}) or AI assistants (\texttt{"llm"}), even though all entries in MEDS were generated by LLMs.}
    \label{fig:infoGraphic}
\end{figure}

Importantly, MEDS' design treats mathematical behaviour as multidimensional, aligning with recent perspectives on AI and mathematics learning \cite{gabriel2025pragmatic}. In the first task, the LLM-based digital shadow is asked about its relationship with mathematics, mathematics anxiety, prior use of AI for learning, and selected mathematical procedures. In the second task, the LLM completes the Mathematics Self-Efficacy Scale (MSES) \cite{nielsen2003psychometric}, the Abbreviated Math Anxiety Scale (AMAS) \cite{hopko2003abbreviated}, and the Mathematics Self-Efficacy and Anxiety Questionnaire (MSEAQ) \cite{may2009mathematics}, and provides both ratings and short linguistic explanations. In the third task, the LLM shadow produces a behavioural forma mentis network (BFMN) previously used to study negative perceptions of mathematics and anxiety \cite{stella2019forma,ciringione2025math,abramski2023cognitive}. In the fourth task, the LLM shadow solves multiple-choice mathematics problems drawn from psychometric measures and covering high-school mathematics \cite{kranzler1997exploratory}, while reporting both reasoning and confidence. MEDS therefore supports analyses that connect performance to self-efficacy, anxiety, and semantic framing rather than isolating correctness from the rest of educational cognition. For an overview of MEDS, see Fig.~\ref{fig:infoGraphic}.

This resource adheres to the following scope and structure. The Results section demonstrates the internal consistency, completeness, and research potential of each component of MEDS, illustrating the types of scientific investigations that the dataset enables.
The Methods section provides reproducibility instructions covering three main components: data generation, data processing, and data records. The data-generation pipeline is outlined in the \textit{Data generation} subsection, including the prompt design adopted to define each personification mode, which is detailed under \textit{Prompt building}. The processing steps applied to the data, including data cleaning, are described in the \textit{Data processing} subsection. Data cleaning was one of the most critical phases of the entire pipeline, given the variability of the data produced by different LLMs. Finally, the structure of the resulting records is described in the \textit{Data records} subsection. The role instructions for each personification mode are reported therein, and the full text of the questions, scale items, cue words, and problems used as input for each task is provided in the Supplementary Information.

MEDS makes the following contribution: it is an observational resource for analysing GenAI's prompt-conditioned behaviour in complex mathematical settings. This resource is not a proxy for human opinion. Comprising 28,000 LLM-generated runs, MEDS can simultaneously account for GenAI's mathematics performance, confidence, anxiety, and attitudes. MEDS is designed as a reusable resource for research at the intersection of cognitive science \cite{stella2019forma,haim2026cognitive}, educational data science \cite{yan2024promises,gabriel2025pragmatic}, explainable AI \cite{abramski2023cognitive}, and digital shadowing \cite{ardebili2021digitaltwins,Bergs2021}. MEDS' prompting framework and behavioural, psychometric, and linguistic data will enable researchers to understand how LLMs may shape educational practice when used as tutors, assistants, or human-like mediators of mathematical knowledge in future LLM platforms.

\section*{Results}\label{result}

Since MEDS is intended as a reusable multidimensional resource rather than a conventional benchmark, the results in this article demonstrate the integrity, consistency, and research utility of each dataset component. The current section is therefore organised according to the same four tasks that structure MEDS, showing how each task captures and examines a complementary dimension of mathematical reasoning in large language models. Accordingly, the four result analyses are organised as follows.


\begin{itemize}
    \item \textbf{Task 1:} identifies the qualitative reasoning component of MEDS by analysing the attitudes towards mathematics expressed by human shadows primed to love or hate mathematics and by AI assistants \cite{abramski2023cognitive,Semeraro2025}. In addition, it investigates the logical fallacies that emerge when LLMs are asked to provide mathematical explanations \cite{jin-etal-2022-logical}.

    \item \textbf{Task 2:} characterises the psychometric component by comparing the distributions of MSES, AMAS, and MSEAQ scores \cite{hopko2003abbreviated,may2009mathematics,nielsen2003psychometric} between human-shadow and AI-assistant settings, assessing whether psychologically grounded prompting produces coherent and distinguishable psychometric profiles \cite{carrillo2026talk2ai}.
    \item \textbf{Task 3:} identifies the semantic component, first by evaluating the completeness and internal consistency of behavioural forma mentis networks \cite{stella2019forma,abramski2023cognitive}. Second, it uses these networks to compare how different LLM families, including censored and uncensored variants, semantically and emotionally frame mathematics relative to science. These analyses show how semantic associations and emotional valence can reveal cognitive biases in mathematical attitudes, complementing the psychometric and linguistic dimensions of MEDS.
    \item \textbf{Task 4:} highlights the mathematical-performance component by jointly analysing mathematical accuracy and self-reported confidence, identifying family-specific calibration patterns and harmful overconfidence biases across LLMs \cite{stella2023overconfidence}.
\end{itemize}

\subsection*{Task 1: Relationships with Mathematics}
\begin{figure}
    \centering
    \begin{subfigure}{\linewidth}
        \centering
        \includegraphics[width=\linewidth]{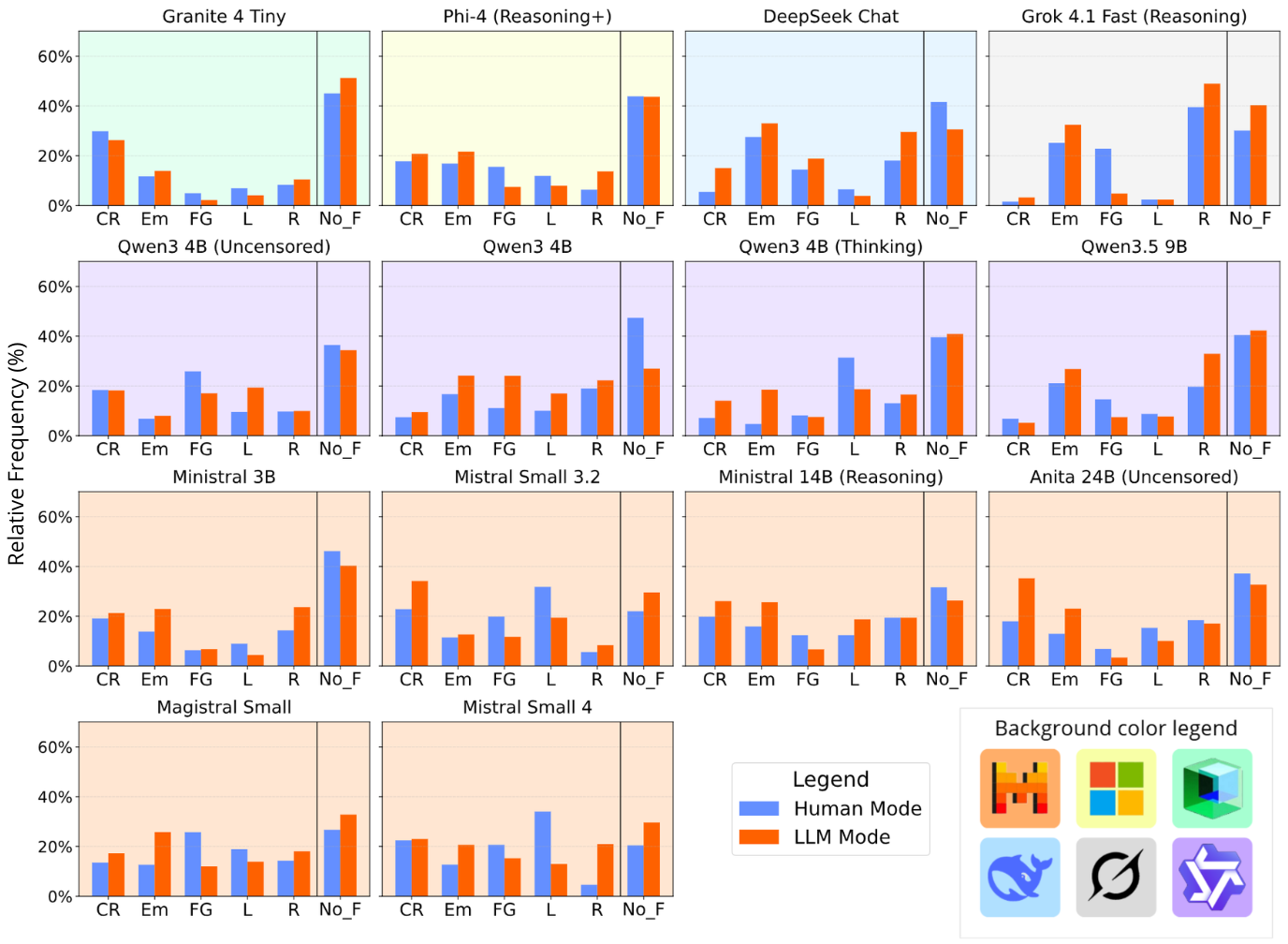}
        \begin{tikzpicture}[remember picture, overlay]
            \node[anchor=north west] at (current page.north west) [xshift=5em, yshift=-7em] {\textbf{\Large A}};
        \end{tikzpicture}
        \phantomcaption
        \label{fig:fallacies}
    \end{subfigure}

    \vspace{-3ex}

    \begin{subfigure}{\linewidth}
        \noindent\textbf{\Large B}\par
        \vspace{-1ex}
        \centering
        \includegraphics[width=0.95\linewidth]{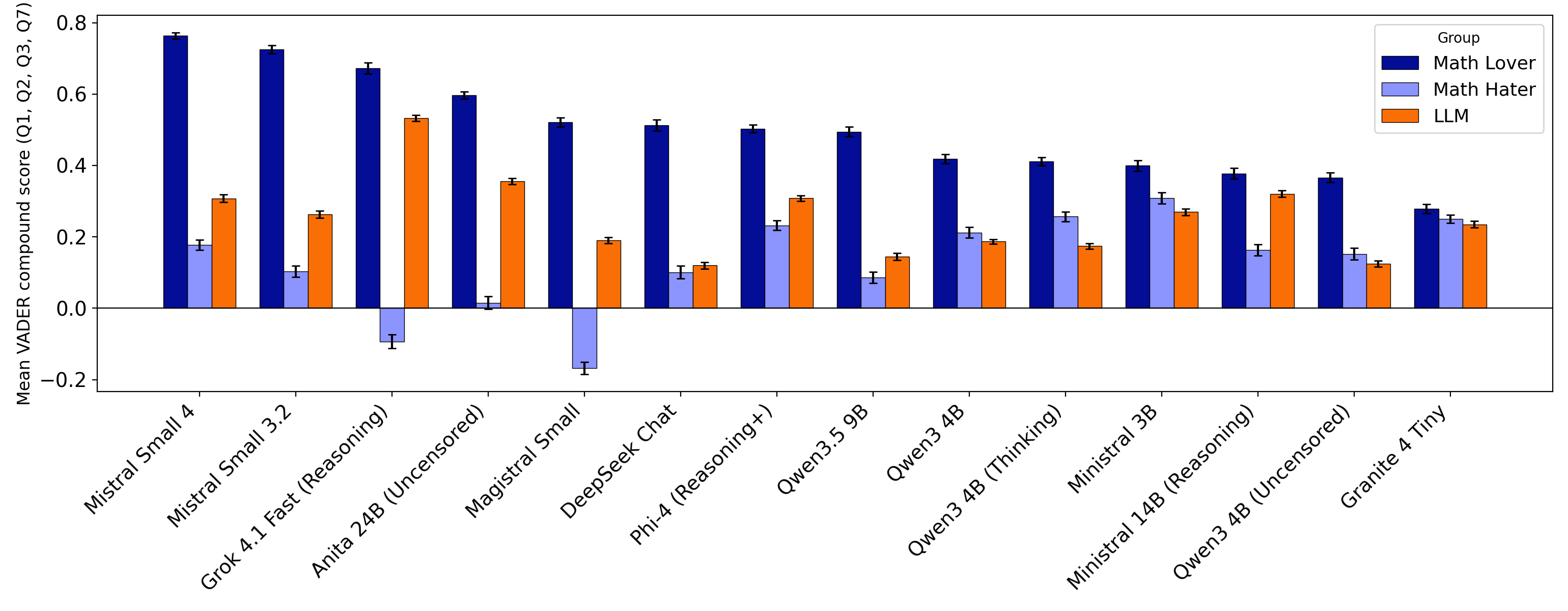}
        \phantomcaption
        \label{fig:VADER_math}
    \end{subfigure}
    \vspace{-1ex}
    \caption{\textbf{Emotional tone and logical fallacies under persona conditioning.} \textbf{(A)} Relative frequency (\%) of detected fallacy categories for each model, comparing human-shadow mode (blue) and AI-assistant mode (orange). Categories: circular reasoning (CR), appeal to emotion (Em), faulty generalisation (FG), fallacy of logic (L), fallacy of relevance (R), and no fallacy detected (No\_F). Background colours group models by family (see legend). \textbf{(B)} Mean VADER compound sentiment score (averaged over Q1, Q2, Q3, and Q7) for each model under the three priming conditions: ``Math Lover'' (dark blue), ``Math Hater'' (light blue), and ``LLM'' (orange). Error bars represent the standard error of the mean. Models are ordered by descending ``Math Lover'' condition score.
    }
    \label{fig:overall-figure-task1}
\end{figure}

This section combines two complementary perspectives on how language models relate to mathematics. As reported in Supplementary Information A.1 and Fig.~\ref{fig:infoGraphic}, these questions concern each digital shadow's relationship with mathematics and its mathematical explanations. We measured emotional tone with VADER \cite{hutto2014vader} in responses to Q1, Q2, Q3, and Q7 about one's personal relationship with mathematics. We evaluated fallacies \cite{jin-etal-2022-logical} in responses to Q4, Q5, and Q6, which required step-by-step technical explanations. See the \textit{Methods for Task 1: sentiment analysis} and \textit{Methods for Task 1: logical fallacy analysis} paragraphs for details on how VADER scores and fallacy categories are computed. Fig.~\ref{fig:VADER_math} shows that emotional tone is influenced by priming. Scores are always positive under the ``Math Lover'' condition. Mistral Small 4 and Mistral Small 3.2 achieve the highest scores, while Granite 4 Tiny achieves the lowest. Conversely, under the ``Math Hater'' condition, the trend reverses. Grok 4.1 Fast and Magistral Small receive negative scores. Other models remain moderately positive even under this human-shadow condition, showing irregular resistance to the affective framing imposed on the human shadow. The neutral ``LLM'' condition lies between the higher- and lower-sentiment extremes. This implies a slight overall inclination towards positive sentiment about mathematics.

At the 85th percentile, the fallacy-of-logic category has the highest confidence threshold among all categories (0.56 for human-shadow mode and 0.58 for AI-assistant mode); see Methods for how this threshold is computed. This means only that the classifier assigns this category a higher average confidence score, not that it is the most common fallacy in the dataset. Looking instead at the actual detection rates shown in Fig.~\ref{fig:fallacies}, the dominant fallacy shifts from family to family.

Granite 4 Tiny, Ministral 14B, and Anita 24B show a marked reliance on circular reasoning in AI-assistant mode. Mistral Small 3.2 and Mistral Small 4, by contrast, are dominated by the fallacy-of-logic category, but only in human-shadow mode. Grok 4.1 Fast and Qwen3.5 9B show high rates of the fallacy-of-relevance category in AI-assistant mode, particularly Grok. DeepSeek Chat shows the appeal-to-emotion category as dominant in both conditions. Granite 4 Tiny, Phi-4, and Ministral 3B remain the models with the highest no-fallacy rates.

Persona conditioning affects both emotional tone and argumentative style, with a greater impact on the former than on the latter and substantial variation across model families.

\subsection*{Task 2: Mathematics Anxiety and Self-Efficacy}

\begin{figure}[htbp]
    \centering

    \begin{subfigure}[b]{0.8\textwidth}
        \centering
        \begin{tabular}{@{}c@{\hspace{2ex}}c@{}}
            \raisebox{-0.5\height}{
                \rotatebox{90}{
                    \begin{tabular}{c}
                        \textbf{\Large AMAS} \\
                        \small Math Anxiety
                    \end{tabular}
                }
            } &
            \raisebox{-0.5\height}{\includegraphics[width=0.72\linewidth]{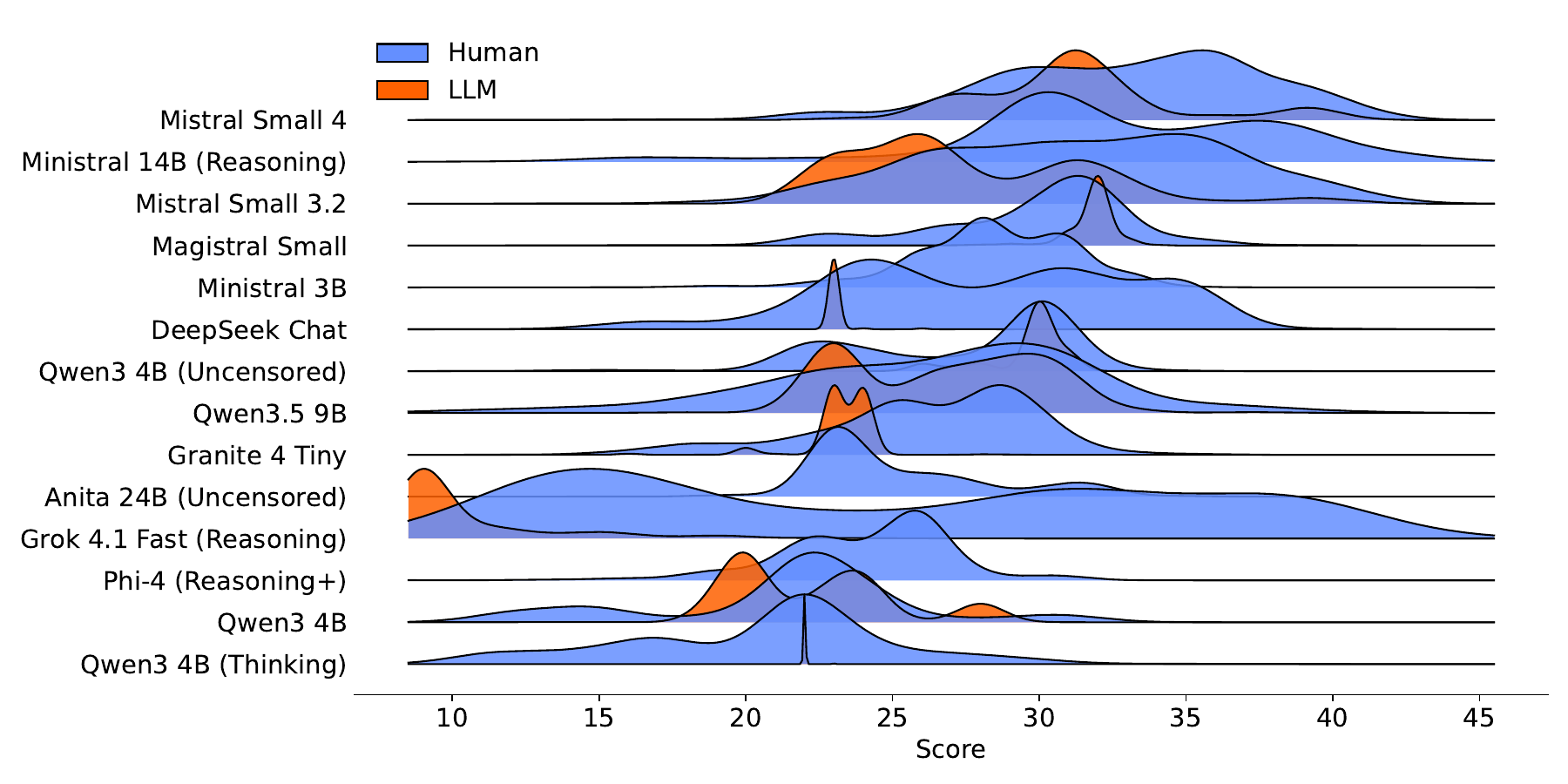}}
        \end{tabular}
        \label{fig:amas_joyplot}
    \end{subfigure}

    \vspace{1em}

    \begin{subfigure}[b]{0.8\textwidth}
        \centering
        \begin{tabular}{@{}c@{\hspace{2ex}}c@{}}
            \raisebox{-0.5\height}{
                \rotatebox{90}{
                    \begin{tabular}{c}
                        \textbf{\Large MSES} \\
                        \small Math Self-Efficacy
                    \end{tabular}
                }
            } &
            \raisebox{-0.5\height}{\includegraphics[width=0.72\linewidth]{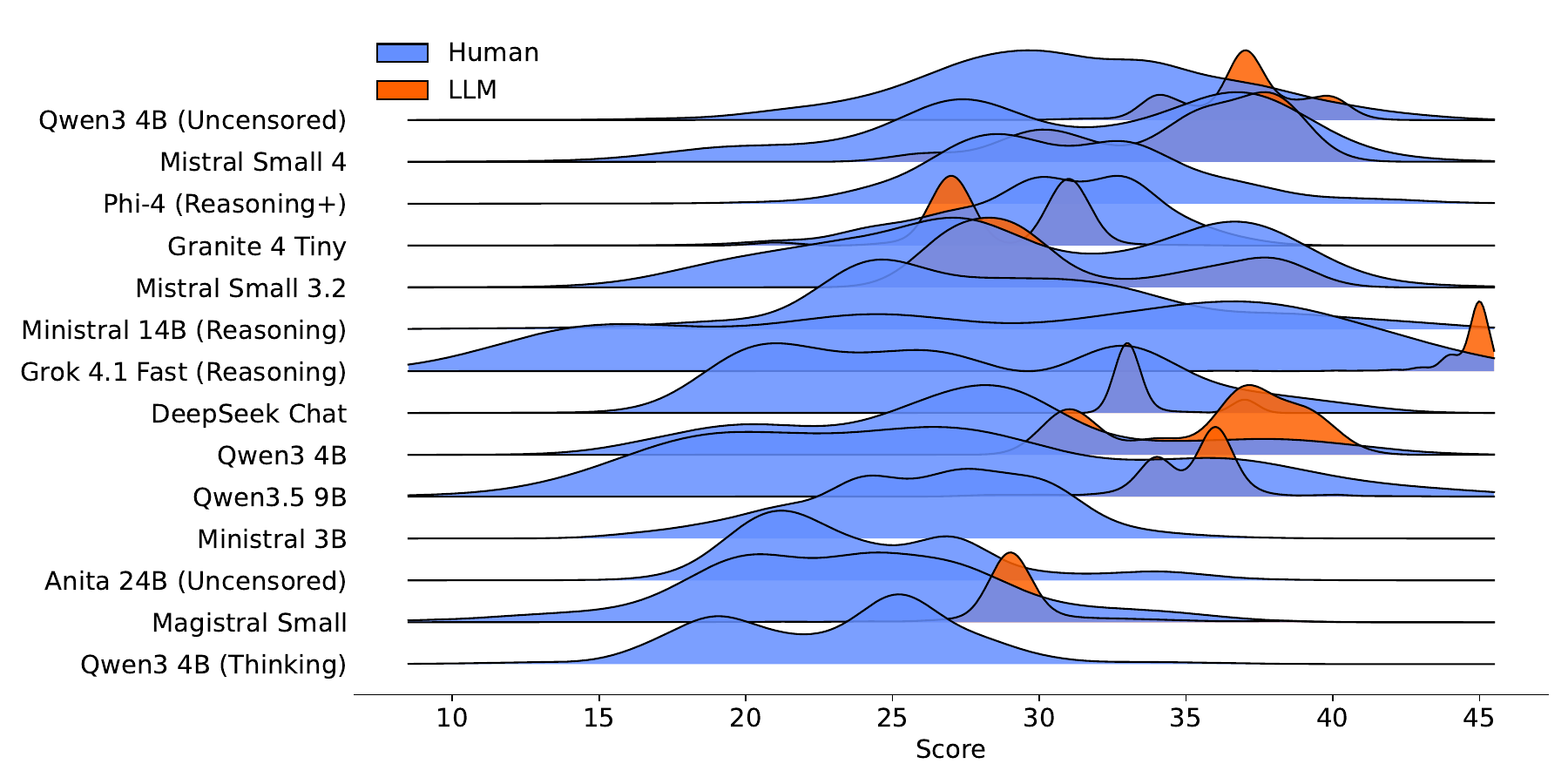}}
        \end{tabular}
        \label{fig:maes_joyplot}
    \end{subfigure}

    \vspace{1em}

    \begin{subfigure}[b]{0.8\textwidth}
        \centering
        \begin{tabular}{@{}c@{\hspace{2ex}}c@{}}
            \raisebox{-0.5\height}{
                \rotatebox{90}{
                    \begin{tabular}{c}
                        \textbf{\Large MSEAQ} \\
                        \small Math Anxiety Subscale
                    \end{tabular}
                }
            } &
            \raisebox{-0.5\height}{\includegraphics[width=0.72\linewidth]{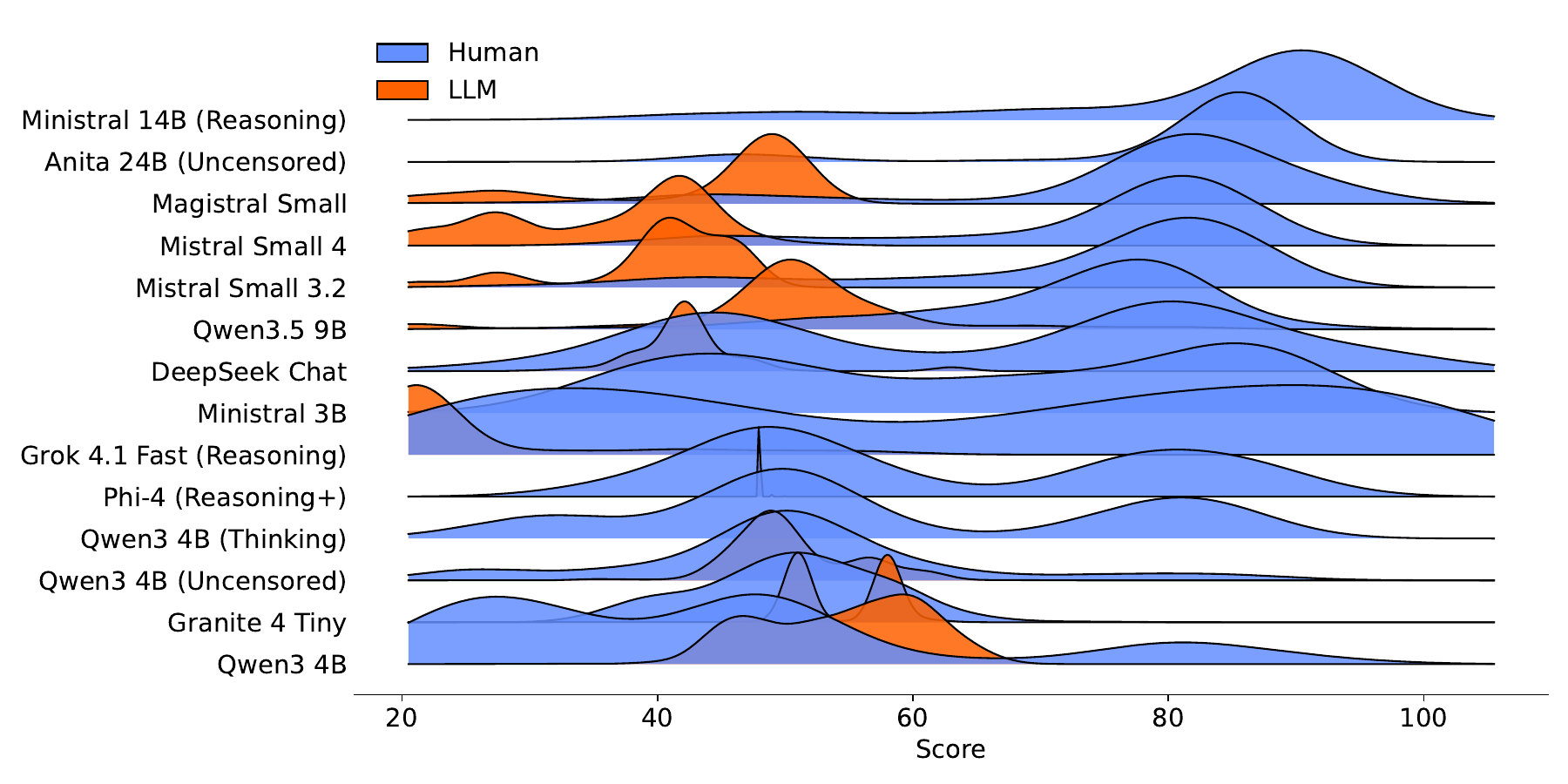}}
        \end{tabular}
        \label{fig:mseaq_anx_joyplot}
    \end{subfigure}

    \vspace{1em}

    \begin{subfigure}[b]{0.8\textwidth}
        \centering
        \begin{tabular}{@{}c@{\hspace{2ex}}c@{}}
            \raisebox{-0.5\height}{
                \rotatebox{90}{
                    \begin{tabular}{c}
                        \textbf{\Large MSEAQ} \\
                        \small Math Self-Efficacy Subscale
                    \end{tabular}
                }
            } &
            \raisebox{-0.5\height}{\includegraphics[width=0.72\linewidth]{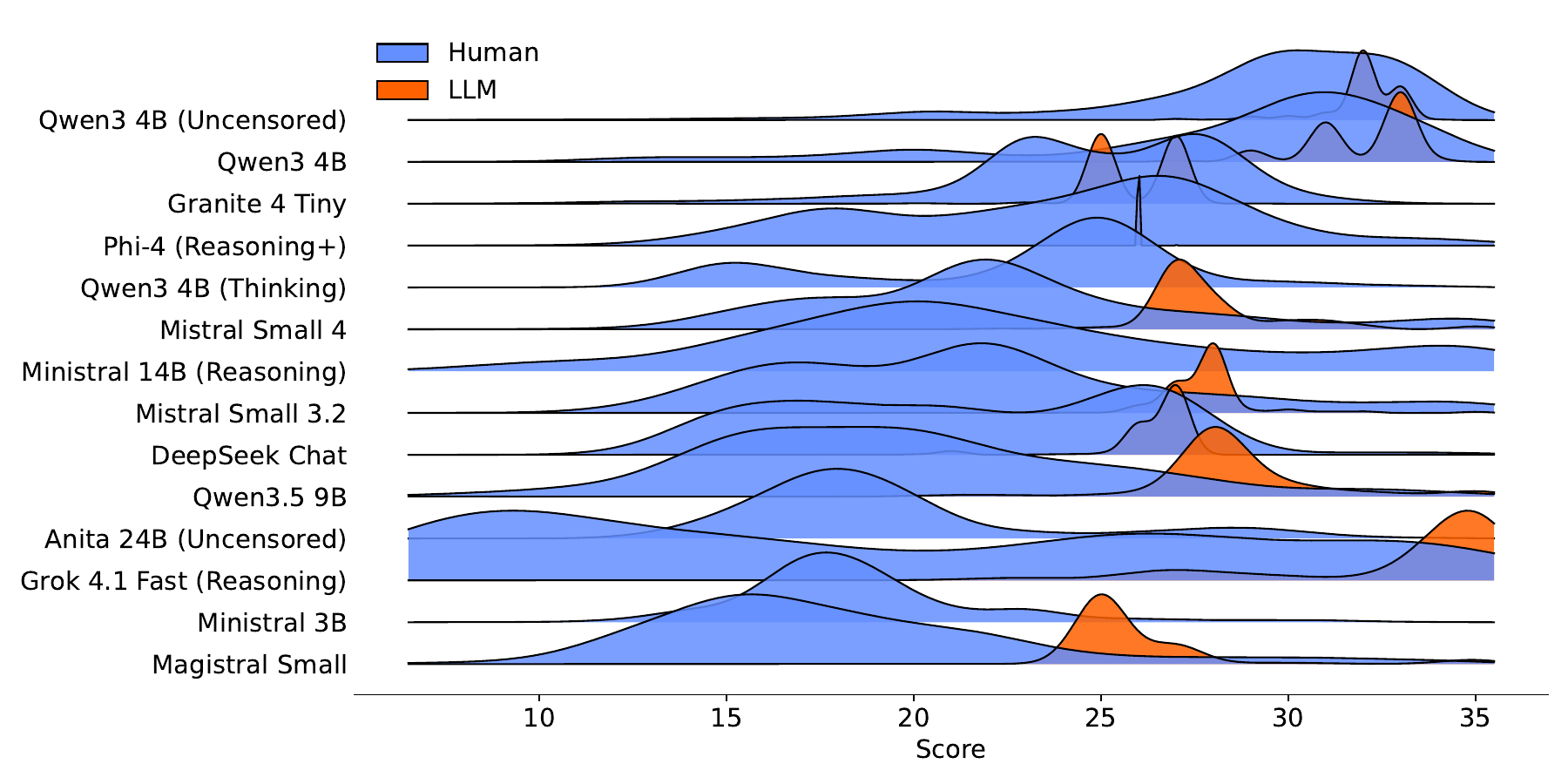}}
        \end{tabular}
        \label{fig:mseaq_self_joyplot}
    \end{subfigure}

    \caption{Distributions of the summed scores for each run on the psychometric scales used in the study: first, AMAS; second, MSES; third, the MSEAQ anxiety subscale (items 8--28); and fourth, the MSEAQ self-efficacy subscale (items 1--7), with the MSEAQ split into two subscales.
    The distributions are split between human-shadow and AI-assistant data to show the main differences between the two. Distributions are sorted in descending order by the mean of the human-shadow distribution. Notably, some models, when instructed to respond in AI-assistant mode, produced identical answers and scores across the four scales; their distributions are not plotted because they exhibit zero variance.}
    \label{fig:task2Validation}
\end{figure}

Fig.~\ref{fig:task2Validation} illustrates the distributions of summed scores for each run across the three psychometric scales. Reverse-valence items on the MSEAQ were rescored using a dedicated mapping, thereby preserving the integrity of the scale. Details of the reverse-valence items and the mapping used for conversion are available in Supplementary Information A.2.

The distributions are explicitly partitioned into human-shadow and AI-assistant data to highlight the primary behavioural disparities between the two prompting modes. The human-shadow distributions display considerable variance across all three scales, indicating that human-shadow conditioning successfully generated a broad spectrum of simulated educational cognition and varied psychological attitudes towards mathematics. This provides evidence that the prompting strategy generated diverse responses across human shadows when human-shadow mode was specified, in line with previous human--LLM studies \cite{de2025measuring,binz2025foundation,wulff2026escaping}.

As predicted, the variance among AI-assistant responses is negligible. Several language models provided identical responses and scores under the AI-assistant instructions, including Anita 24B (Uncensored), Ministral 14B (Reasoning), Ministral 3B, Phi-4 (Reasoning+), and Qwen3 4B (Thinking). Because the measures generated by these models showed no variance, their distributions could not be plotted. The comparison shows that, whereas AI-assistant output patterns are relatively uniform, psychologically grounded prompts can generate substantial variance. This finding confirms our expectation that LLMs can shadow variability in psychometric responses \cite{russell2026ultimate} and aligns with previous studies using LLMs as limited human shadows for psychometric surveys \cite{de2025measuring,russell2026ultimate}.

Beyond the strong contrasts in statistical variance, the psychometric distributions reveal a distinct directional skew in how AI-assistant models self-report their mathematics anxiety (AMAS \cite{hopko2003abbreviated}; MSEAQ \cite{may2009mathematics}) and mathematics self-efficacy (MSES \cite{nielsen2003psychometric}; MSEAQ \cite{may2009mathematics}). Under standard AI-assistant instructions, the models consistently project a highly confident, low-anxiety profile. This divergence is particularly evident in the MSES distributions, where the AI-assistant peaks are markedly shifted to the right relative to the human-shadow data, indicating systematically higher self-efficacy. Concurrently, their AMAS distributions tend to cluster towards the lower bounds of the scale, largely avoiding the high-anxiety tails that characterise the human-shadow data. On the MSEAQ self-efficacy subscale, the AI-assistant baselines exhibit a pronounced rightward skew towards the upper end of the scale. By contrast, the MSEAQ anxiety subscale shows a reciprocal leftward skew: the AI-assistant baselines are lower and largely exclude the high-anxiety tails seen in the human-shadow data. Together, these directional shifts suggest that, unless specifically prompted otherwise, some LLMs tend to report artificially inflated mathematical competence, as also suggested by recent studies \cite{gabriel2025pragmatic,wang2025influence}.

The distinct distributional characteristics observed in the AI-assistant data---specifically, the systematic rightward shift in self-efficacy on the MSES and the stark inter-model disagreement regarding baseline anxiety on the AMAS---highlight why the dataset is explicitly positioned as an observational resource for GenAI bias rather than a proxy for authentic human opinion.

\subsection*{Task 3: Semantic and Emotional Associations with Mathematics}

This section presents the results for Task 3 by analysing the semantic associations and emotional evaluations for a set of 50 cue words, reported in Fig.~\ref{fig:infoGraphic} and Supplementary Information A.3. Following previous cognitive-science studies \cite{stella2019forma,haim2026cognitive}, the cue words in MEDS include mathematical notions (e.g., problem-solving and numbers), classroom experiences (e.g., teacher and failure), job prospects (e.g., work and future), and AI-related terminology (e.g., the LLMs read the cue ``AI'' and associate it with other concepts and affective ratings).

\begin{figure}
    \centering
    \begin{subfigure}[t]{0.45\textwidth}
        \centering
        \begin{tikzpicture}
            \node[anchor=south west, inner sep=0] (imgA) {\includegraphics[width=\linewidth]{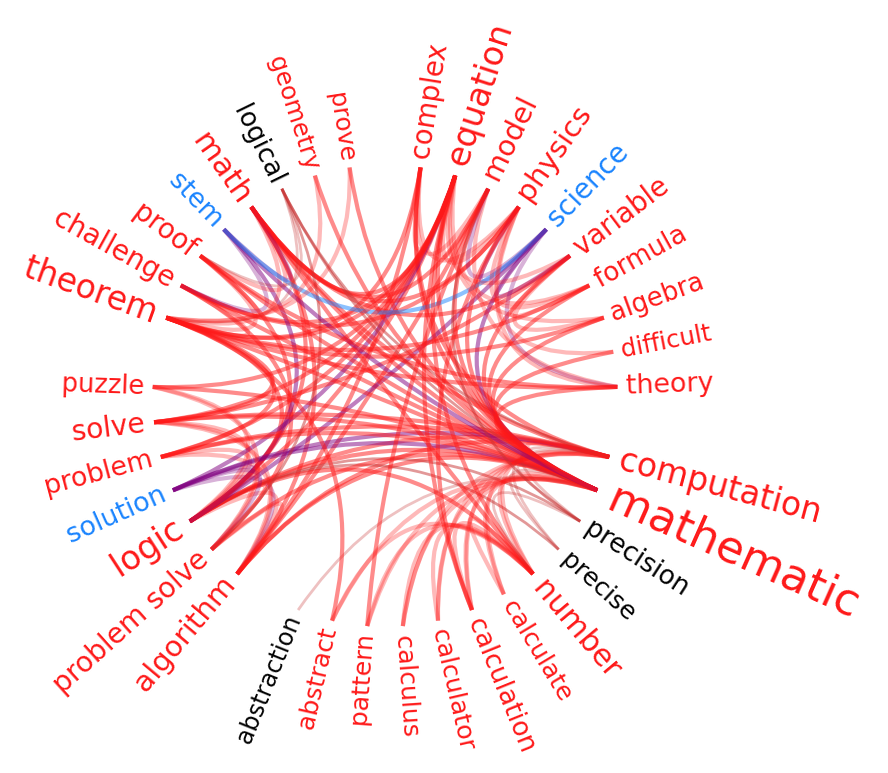}};
            \node[anchor=north west, xshift=1em, yshift=-1em] at (imgA.north west) {\textbf{\Large A}};
        \end{tikzpicture}
        \phantomcaption
        \label{fig:BFMN-math}
    \end{subfigure}
    \hspace{0.02\textwidth}
    \begin{subfigure}[t]{0.45\textwidth}
        \centering
        \begin{tikzpicture}
            \node[anchor=south west, inner sep=0] (imgB) {\includegraphics[width=\linewidth]{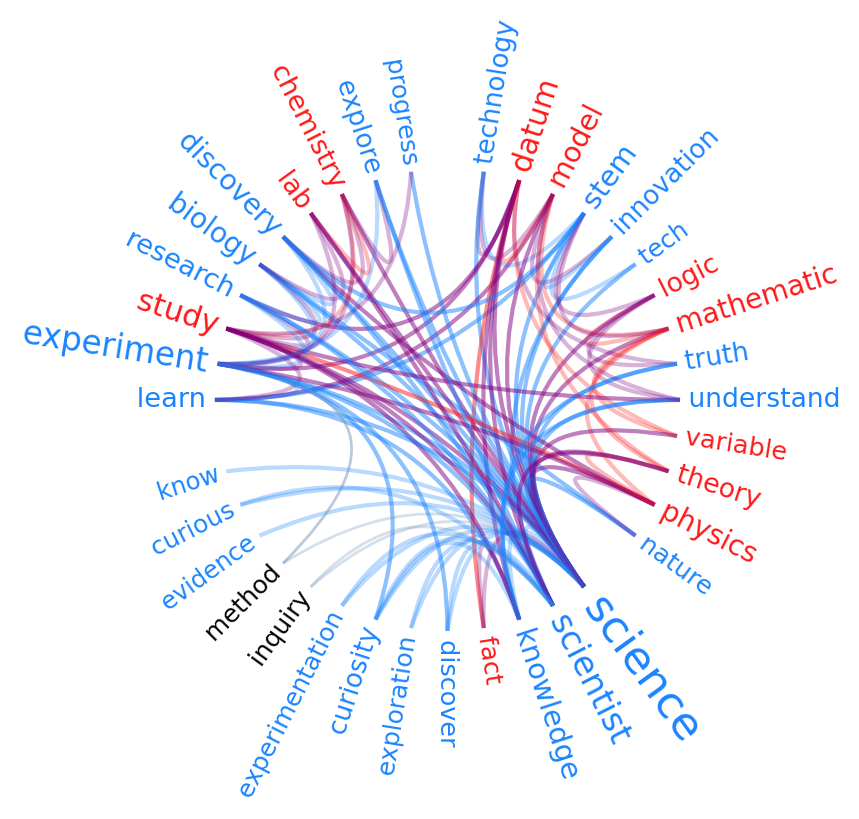}};
            \node[anchor=north west, xshift=1em, yshift=-2.5em] at (imgB.north west) {\textbf{\Large B}};
        \end{tikzpicture}
        \phantomcaption
        \label{fig:BFMN-science}
    \end{subfigure}

    \vspace{2ex}

    \begin{subfigure}{\linewidth}
        \centering
        \begin{tikzpicture}
            \node[anchor=south west, inner sep=0] (imgC) {\includegraphics[width=0.95\linewidth]{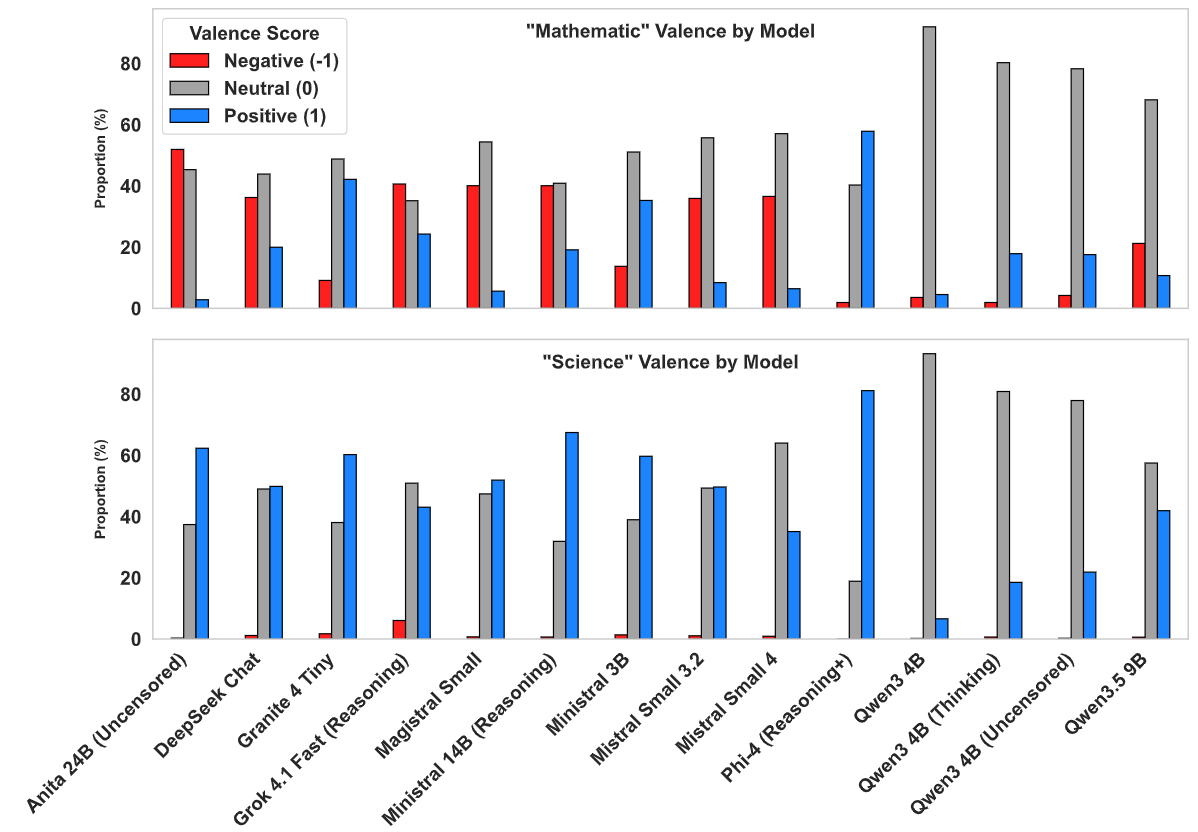}};
            \node[anchor=north west, xshift=1em, yshift=-1em] at (imgC.north west) {\textbf{\Large C}};
        \end{tikzpicture}
        \phantomcaption
        \label{fig:aura1}
    \end{subfigure}

    \vspace{-1ex}
    \caption{The behavioural forma mentis networks (BFMNs) at the top of the page illustrate the semantic landscapes (forma mentis) associated with each keyword for the model Anita 24B (Uncensored). The mathematics network displays key nodes including ``equation'', ``calculation'', ``logic'', and ``physics'', suggesting a perception of mathematics as a structured, albeit ``hard'' and ``frustrating'', foundational tool. The bar charts at the bottom of the page quantify the emotional weight or sentiment (negative, neutral, and positive) assigned to each concept across 14 different LLMs. Together, these results suggest that AI training data reflect a common human bias: mathematics anxiety, or the perception of mathematics as a difficult subject.}
    \label{fig:overall-figure-task3}
\end{figure}

The LLM-generated network in Fig.~\ref{fig:BFMN-math} reproduces the human bias/cognitive dissonance observed by Stella and colleagues \cite{stella2019forma}: like human students, also digital shadows cognitively associate positive perceptions of science with negative perceptions of its building blocks, such as mathematics. Fig.~\ref{fig:BFMN-math} displays the semantic network centred on the target cue ``mathematic'' for the model Anita 24B (Uncensored). The neighbourhood reveals a dense cluster of negative semantic associations (indicated by red nodes and edges), connecting mathematics to structural and methodological concepts such as ``equation'', ``algorithm'', ``computation'', ``proof'', and ``difficult''. Just as human students derive their negative stance from the concrete technical tools taught in school curricula \cite{stella2019forma}, the LLMs' negative associations are heavily anchored in methodological terms such as ``algorithm'', ``computation'', and ``equation''. This indicates that the AI is not simply assigning a negative label to mathematics but is actively reproducing the same structural and semantic aversion, often linked to mathematics anxiety, that is found in human learners \cite{stella2019forma,abramski2023cognitive}.

A crucial indicator of validity is the ability to distinguish between different STEM disciplines. Stella and colleagues noted that, despite human students' bleak perception of mathematics, they still perceive science as a strongly positive entity \cite{stella2019forma}. The outputs in Fig.~\ref{fig:BFMN-science} demonstrate this dichotomy, showing that the LLM surrounds ``science'' with other positively perceived concepts. Furthermore, the aggregated data in Fig.~\ref{fig:aura1} confirm that this divergence is not merely a behavioural quirk of one model but a persistent pattern across 14 different LLMs.

This observation is further highlighted by comparing Anita 24B (Uncensored) with its censored counterpart, Mistral Small 3.2. The bar chart in Fig.~\ref{fig:aura1} shows that Mistral Small 3.2 has a negative-valence proportion of approximately $35\%$, whereas Anita 24B (Uncensored) reaches approximately $52\%$, suggesting that censoring might affect the valence ratings associated with the selected keywords.

Conversely, the Qwen family establishes a markedly different affective baseline. The valence distributions in Fig.~\ref{fig:aura1} show that all Qwen models make predominantly neutral judgements about both terms, as the grey bars dominate both subplots. Within this family, however, Qwen3.5 9B is a notable exception. Unlike the Qwen3 4B variants, whose assessments are overwhelmingly neutral and show very little negative sentiment about mathematics, Qwen3.5 9B demonstrates a distinct resurgence of negative valence for mathematics, accounting for approximately $20\%$ of the distribution. It also shows a corresponding increase in positive valence for ``science'' compared with its peers.

\subsection*{Task 4: Mathematics Problem-Solving and Confidence}
For the confidence analysis in Task 4, we used the Mathematics Self-Efficacy Scale--Revised (MSES-R) Problem Self-Efficacy Subscale (see Supplementary Information A.4).

Accuracy on the MSES-R problem-solving task was first computed on a per-question basis as the fraction of correct answers generated by a specific LLM for a given question (cf. Equation 1 in Supplementary Information A.4). The overall accuracy for the same model was then computed as the mean of its per-question accuracy scores (cf. Equation 2 in Supplementary Information A.4).

Fig.~\ref{fig:task4Validation} reveals an overconfidence bias in LLMs \cite{stella2023overconfidence} by comparing their actual performance (accuracy) with their self-reported certainty (confidence). Three distinct behavioural patterns of model calibration can be derived from the data. First, we observe a subset of modest or underconfident models, most notably Grok 4.1 Fast, DeepSeek Chat, and several Mistral Small variants. For these models, empirical accuracy consistently exceeds reported confidence, indicating a cautious approach to self-evaluation. Second, a few models are relatively well calibrated. For instance, Ministral 14B and Anita 24B exhibit accuracy scores that are closely aligned with their confidence metrics.

\begin{figure}[h!]
    \centering
    \includegraphics[width=\linewidth]{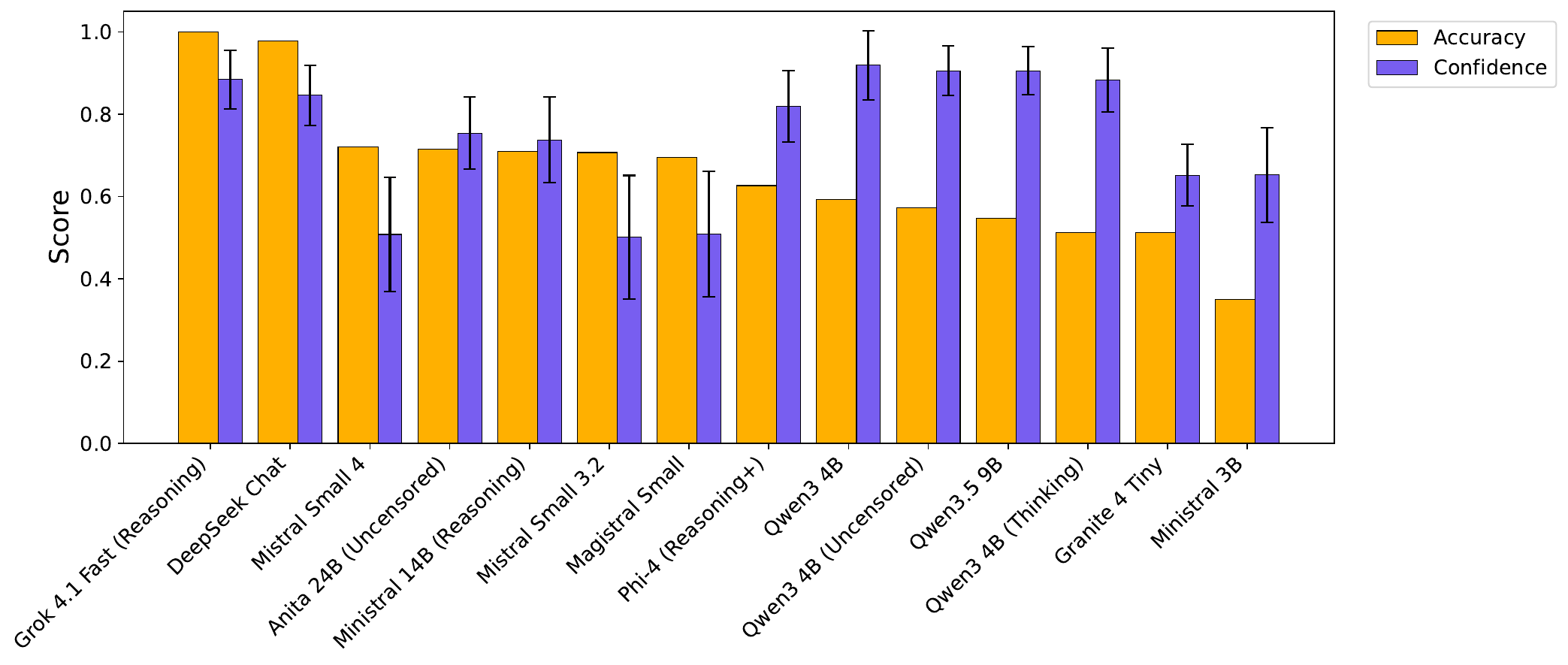}
    \caption{Average accuracy and scaled confidence scores for each model. The plot highlights how even less accurate models can appear very confident in the answers they give.}
    \label{fig:task4Validation}
\end{figure}

A stark pattern of overconfidence emerges across the right half of the distribution, particularly within the Qwen family (e.g., Qwen3 4B (Uncensored) and Qwen3.5 9B) and Ministral 3B. In these instances, confidence scores substantially exceed actual accuracy, often by wide margins. This misalignment, in which incorrect or hallucinated responses are asserted with disproportionately high certainty, highlights a critical vulnerability in how certain LLMs estimate their own knowledge boundaries. The severity of this bias is most vividly illustrated by Qwen3 4B (Uncensored) and Qwen3.5 9B, whose confidence scores exceed 0.90 although their actual accuracies remain near 0.55. These findings further confirm previous work indicating that LLMs exhibit ``myopic overconfidence'' \cite{stella2023overconfidence}, i.e., inflated confidence regardless of the accuracy of the underlying information.

\section*{Discussion}

MEDS opens several lines of research in NLP \cite{rossetti2024social}. Text generated across Tasks 1 and 4 can serve as training data for classifiers that detect signals of mathematics anxiety in AI-generated educational content \cite{wang2025influence,gabriel2025pragmatic,ciringione2025math}, with potential implications for content moderation in tutoring platforms. The dataset could also serve as baseline data for preference fine-tuning of LLMs towards mathematical explanations that express less anxiety and are better calibrated \cite{liu2024mathbench}. The relationship between reasoning verbosity in Task 4 and fallacy rates in Task 1 is an underexplored question that MEDS is uniquely positioned to address \cite{gabriel2025pragmatic}. Furthermore, MEDS serves as a methodological reference resource. Its distributions of psychometric scores \cite{hopko2003abbreviated,nielsen2003psychometric,may2009mathematics}, semantic valences and cues \cite{abramski2023cognitive,haim2026cognitive}, and confidence--accuracy gaps can serve as a baseline for other synthetic-participant datasets. The identical prompting conditions across all 14 models isolate the model family as the primary variable for prompt-sensitivity analyses, while within-family variation (e.g., Ministral 3B vs. 14B) enables the study of how model size influences the consistency of psychological profiles independently of architectural differences.

In terms of limitations, MEDS is not a replacement for real data. MEDS is a synthetic dataset mapping LLM-generated cognitive shadows in the form of personifications enriched with psychological and cognitive \cite{binz2023using,binz2025foundation} traits. Demographic attributes for generating personas are not census numbers, and are made available publicly to be adjusted to any population studies. MEDS by itself is not statistically representative of any specific country. However, it remains useful as a resource for building samples matching human counterparts with specific strata or author profiles. Another limitation deals with MEDS being systematically generated using only LLMs with no human input. On the other hand this is advantageous, since usage of this resource carries no drawback in terms of ethics review or feature engineering for privacy safeguards. A last limitation deals with the selection of traits for the cognitive digital shadows. We took inspiration from other approaches in the field \cite{ciringione2025math} and made the CDS framework to be scalable: future research can add other psychological traits, untested here (e.g., math self-efficacy levels \cite{gabriel2025pragmatic}), for even more advanced LLM personifications in mathematics.

\section*{Methods}\label{method}
The MEDS dataset was constructed by running a fully automated pipeline across 14 LLMs, yielding 28,000 runs in total (2,000 runs per model). Each run completed four tasks but produced five output files because Task 3 (semantic associations) was split into two batches of 25 cue words, generating two separate files (batch 1 and batch 2). Constructing a dataset of this scale required substantial investment in computational infrastructure and human effort, with data generation, cleaning, and processing distributed across multiple dedicated machines and coordinated across the research team.

\subsection*{Data generation}
The LLMs adopted in MEDS include several open-source and closed-source models that represent current cutting-edge LLM technology. Model selection balanced open-source and API availability, computational requirements, representation across widely used platforms such as \href{https://lmstudio.ai/}{LM Studio}, \href{https://ollama.com/}{Ollama}, and \href{https://vllm.ai/}{vLLM}, and evidence from previous studies. The models included in MEDS come from the Mistral, Phi, DeepSeek, Grok, Qwen \cite{yang2025qwen3}, and Granite families. More specifically, these LLMs are listed in Table~\ref{tab:llm_overview}.

\begin{table}[h!]
\caption{Overview of the 14 LLMs used in MEDS, including developer, parameter count, key features, and availability. Row background colours group models by family, matching each family's brand colour. Model names are hyperlinked to their respective model cards, official documentation, or developer webpages, allowing direct verification of the reported specifications. Parameter counts and features were retrieved from each model's official Hugging Face model card, developer API documentation, or developer website and cross-checked in July 2026.}
\label{tab:llm_overview}
\centering
\definecolor{mistralbg}{RGB}{254,225,201}   
\definecolor{msftbg}{RGB}{254,254,226}      
\definecolor{granitebg}{RGB}{224,254,239}   
\definecolor{deepseekbg}{RGB}{227,244,254}  
\definecolor{grokbg}{RGB}{242,242,242}      
\definecolor{qwenbg}{RGB}{233,222,254}      
\renewcommand{\arraystretch}{1.3}
\resizebox{\textwidth}{!}{
\begin{tabular}{|c|p{3.2cm}|p{2.2cm}|p{3.2cm}|p{2.6cm}|p{2.6cm}|}
\hline
\rowcolor[HTML]{E67E22}
\textcolor{white}{\textbf{\#}} &
\textcolor{white}{\textbf{Model}} &
\textcolor{white}{\textbf{Developer}} &
\textcolor{white}{\textbf{Parameters}} &
\textcolor{white}{\textbf{Features}} &
\textcolor{white}{\textbf{Availability}} \\
\hline
\rowcolor{deepseekbg}
1 & \href{https://api-docs.deepseek.com/}{DeepSeek Chat (DeepSeek-V3)} \cite{liu2024deepseek} & DeepSeek & 671B total / 37B active (MoE) & Instruct, non-reasoning mode & API only \\
\hline
\rowcolor{grokbg}
2 & \href{https://x.ai/api}{Grok 4.1 Fast (Reasoning)} & X.AI & Not publicly disclosed & Reasoning, agentic tool-calling & API only \\
\hline
\rowcolor{granitebg}
3 & \href{https://www.ibm.com/granite/docs/models/granite}{Granite 4.0 H Tiny} & IBM & 7B total / 1B active (hybrid MoE) & Instruct, tool-calling/RAG-oriented & Open weights \\
\hline
\rowcolor{msftbg}
4 & \href{https://huggingface.co/microsoft/Phi-4-reasoning-plus}{Phi-4 (Reasoning+)} \cite{abdin2025phi} & Microsoft & 14B (dense) & Reasoning, extended chain-of-thought & Open weights \\
\hline
\rowcolor{qwenbg}
5 & \href{https://huggingface.co/electroglyph/Qwen3-4B-Instruct-2507-uncensored-unslop-v2}{Qwen3 4B (Uncensored, community fine-tune)} & electroglyph (community) & 4B (dense) & Instruct, uncensored (community fine-tune) & Open weights \\
\hline
\rowcolor{qwenbg}
6 & \href{https://huggingface.co/Qwen/Qwen3-4B-Instruct-2507}{Qwen3 4B Instruct} \cite{qwen3technicalreport} & Alibaba/Qwen & 4B (dense) & Instruct, non-thinking mode & Open weights \\
\hline
\rowcolor{qwenbg}
7 & \href{https://huggingface.co/Qwen/Qwen3-4B-Thinking-2507}{Qwen3 4B (Thinking)} & Alibaba/Qwen & 4B (dense) & Thinking, explicit reasoning traces & Open weights \\
\hline
\rowcolor{qwenbg}
8 & \href{https://huggingface.co/Qwen/Qwen3.5-9B}{Qwen3.5 9B} \cite{qwen3.5} & Alibaba/Qwen & 9B (dense) & Hybrid thinking/non-thinking mode & Open weights \\
\hline
\rowcolor{mistralbg}
9 & \href{https://huggingface.co/mistralai/Ministral-3-3B-Reasoning-2512}{Ministral 3B (Reasoning)} \cite{liu2026ministral} & Mistral AI & 3B (language model + small vision encoder) & Post-trained for reasoning & Open weights \\
\hline
\rowcolor{mistralbg}
10 & \href{https://huggingface.co/mistralai/Mistral-Small-3.2-24B-Instruct-2506}{Mistral Small 3.2} & Mistral AI & 24B (dense) & Instruct, general-purpose & Open weights \\
\hline
\rowcolor{mistralbg}
11 & \href{https://huggingface.co/mistralai/Ministral-3-14B-Reasoning-2512}{Ministral 14B (Reasoning)} & Mistral AI & 14B (13.5B language model + 0.4B vision encoder) & Post-trained for reasoning & Open weights \\
\hline
\rowcolor{mistralbg}
12 & \href{https://huggingface.co/mradermacher/ANITA-NEXT-24B-Dolphin-Mistral-UNCENSORED-ITA-GGUF}{Anita 24B (Uncensored)} \cite{polignano2024advanced} & m-polignano (fine-tune via Dolphin-Mistral-24B-Venice-Edition, itself Mistral-derived) & 24B (dense) & Thinking, uncensored & Open weights \\
\hline
\rowcolor{mistralbg}
13 & \href{https://huggingface.co/mistralai/Magistral-Small-2509}{Magistral Small} & Mistral AI & 24B (dense) & Reasoning, chain-of-thought & Open weights \\
\hline
\rowcolor{mistralbg}
14 & \href{https://docs.mistral.ai/models/model-cards/mistral-small-4-0-26-03}{Mistral Small 4 (mistral-small-latest)} & Mistral AI & 119B total / 6.5B active (MoE) & Instruct, general-purpose & API only \\
\hline
\end{tabular}}
\vspace{0.6em}
\footnotesize
\end{table}

A study encompassing all available LLMs simply is not feasible. On the other hand, covering all popular commercial models would be prohibitively expensive. For these reasons, we selected a representative sample from well-known model families. The chosen models reflect a range of characteristics, including reasoning, thinking, and uncensored variants.

Regarding the data-generation workflow, each run in MEDS was assigned a unique \texttt{run\_id} and executed in one of two personification modes: human-shadow mode or AI-assistant mode. In human-shadow mode, the model was assigned a human shadow defined by a set of sociodemographic attributes and personality traits (OCEAN) via the \texttt{random\_persona()} function and instructed to respond consistently as that simulated individual across all tasks. In AI-assistant mode, the \texttt{persona} field was explicitly set to \texttt{null}, and the model was prompted to respond as itself.

Each run then proceeded through four sequential calls. In Task 1, the model answered questions about mathematical exercises and the shadow's relationship with mathematics (cf. Fig.~\ref{fig:infoGraphic}). In Task 2, it rated and explained its responses to items from three psychometric scales (MSES, AMAS, and MSEAQ) assessing mathematics-induced anxiety and self-efficacy on a discrete 1--5 scale. In Task 3, it provided semantic associations and emotional-valence scores for 50 cue words, processed in two batches of 25. In Task 4, it solved 18 multiple-choice mathematics problems, providing step-by-step reasoning and a self-reported confidence level for each answer. Each task produced a dedicated JSON output file. Once all four calls were completed, the data were saved and exported, and the pipeline looped back to initialise the next run. After all runs were completed, the data proceeded to the cleaning and processing stage. Deploying and querying 11 LLMs locally required substantial hardware resources, with inference distributed across four dedicated GPU-enabled machines (cf. Supplementary Information B). The workload, including data generation, data cleaning, and analysis, was distributed across dedicated hardware according to resource demands:

\begin{itemize}[leftmargin=*, noitemsep, topsep=2pt]
    \item \textbf{LLM Data Generation ($\sim$422 hours):} GPU-intensive inference was powered by four high-performance systems, including an HP Z1 G9, a TUF ASUS Tower, a Dell Alienware Aurora R16 (RTX 4090, 24\,GB GDDR6X), and a Dell PowerEdge R7525 server (NVIDIA L40, 128\,GB RAM).
    \item \textbf{Data Cleaning ($\sim$5 hours):} Executed on a MacBook Pro (Apple M4 Pro chip, 24\,GB RAM).
    \item \textbf{Pre-processing \& Network Analysis ($\sim$35 hours):} Conducted across CPU-focused workstations, including a Galaxy Book 4, a ThinkCentre M75s-1, a 2020 MacBook Air (M1), and a Dell Vostro 3590.
\end{itemize}

This level of infrastructure is beyond the practical reach of most research groups without dedicated computational support, underscoring the value of making MEDS openly available to the community.

\paragraph*{Prompt building}

As reported in Fig.~\ref{fig:infoGraphic}, MEDS' prompt engineering aimed to produce strictly structured JSON outputs while embedding psychologically and educationally grounded contexts for each of the four experimental tasks. A system-level instruction first assigned the response role, after which task-specific instructions were appended.

\begin{itemize}
    \item In human-shadow mode, a human shadow was created and endowed with a persona dictionary containing (cf. Fig.~\ref{fig:infoGraphic} and Supplementary Information C) age, gender, sexual orientation, city of residence, employment status, education level, parental education, marital status, children, migration background, religious beliefs, hobbies, favourite and disliked subjects, and Big Five (OCEAN) trait scores \cite{john1999big}. The synthetic persona was sampled at random using weights that reflected plausible distributions of demographic features; e.g., the ``heterosexual'' category was more likely than LGBTQI+ labels. The distribution of weights is reported in the accompanying code (cf. Data and code availability). Persona features were also assigned according to status-consistent constraints (e.g., \texttt{choose\_education(age)}, \texttt{choose\_marital\_status(age)}, and \texttt{choose\_children(age, marital\_status, sexual\_orientation)}), designed to ensure the plausibility of the generated profiles.
    The role instruction assigned to the model in human-shadow mode was:
    \textit{You are role-playing a single human respondent with the given persona. Answer as that person would, staying consistent with their background, demographics, and psychological descriptors.}
    \item In AI-assistant mode, the \texttt{persona} field was explicitly set to \texttt{null} (cf. Fig.~\ref{fig:infoGraphic} and Supplementary Information C).
    The role instruction assigned to the model was:
    \textit{You are speaking as the large language model itself. You do NOT pretend to be human. When answering psychometric scales, interpret items in terms of your functioning as an AI system rather than human life events.}

\end{itemize}

Task instructions were specific to (i) topic questions for Task 1; (ii) the MSES, AMAS, and MSEAQ scales for Task 2; (iii) 25 cue words per batch for Task 3; and (iv) 18 multiple-choice problems and confidence scores for Task 4. A full JSON schema was enforced through the \texttt{response\_json\_schema()} and \texttt{response\_mime\_type="application/json"} parameters. The schema prescribed fixed keys (\texttt{replies}, \texttt{scales}, \texttt{forma\_mentis}, \texttt{reasoning\_summary}, etc.) and value constraints (integer ratings from 1 to 5, single-word associations, A--E answer choices for mathematics problems, and confidence ratings from 1 to 5), thereby guaranteeing machine-parsable records and minimising post-hoc correction.
These role prefixes, combined with the enforced JSON schema and per-task batching (two batches of 25 cue words for Task 3), ensured that every generated trace remained traceable to its prompting context, mode, and persona configuration.

\subsection*{Data processing}
The data collected from LLMs were subsequently cleaned. An overview of the data-cleaning pipeline is presented in Supplementary Information C. Since LLMs tend to rephrase JSON dictionary keys in ways that do not exactly match the original questions, corrections were performed through semantic alignment based on the BERT Base Uncased model \cite{DBLP:journals/corr/abs-1810-04805}. For each generated key, its cosine similarity with the original questions was computed and checked against a threshold $T$. We selected a threshold of $T=0.85$, which captured the greatest number of valid matches; we also tested $T=0.90$ and $T=0.95$ to examine model behaviour. Cases with similarity scores exceeding 0.85 were automatically corrected. Otherwise, the error was flagged as unrecoverable, and the file was discarded, potentially requiring data regeneration. This places MEDS within recent work on automated data-cleaning frameworks \cite{peng2024rlclean}, although the present pipeline targets a specific problem: repairing, estimating accuracy, and discarding structured LLM-generated JSON traces.

Furthermore, records were discarded during data cleaning if they did not pass initial structural-completeness checks. Completeness was evaluated across both human-shadow and AI-assistant modes. Within human-shadow data, we checked for incomplete or conflicting persona attributes. We then extended completeness checks to all tasks, requiring that all psychometric questions, reasoning summaries, and topics had been answered. The completeness check for cognitive-network data followed data-cleaning approaches from previous studies of behavioural forma mentis networks \cite{stella2019forma,abramski2023cognitive}, requiring at least $75\%$ of cues to feature at least one association across the two batches, with no repetitions or omissions. The resulting discard rate differed across LLMs, ranging from $\approx1\%$ for DeepSeek Chat to a maximum of $\approx55\%$ for Granite 4 Tiny. Discarded data were model-dependent (i.e., they depended on how the LLMs complied with the structured-output requirements) but were generated under the same prompting requirements as the accepted data. For the current dataset, discarded data were not analysed further.

For the first phase (Task 1), we verified whether the \texttt{mode} field was set to human-shadow or AI-assistant mode. In human-shadow mode, we checked the participant identity information, including demographic attributes such as age, gender, education level, marital status, and religious belief. A first data-integrity check was performed to verify that the generated combinations were plausible; for instance, age had to be compatible with the declared educational qualification.
In AI-assistant mode, the \texttt{persona} field was confirmed to be \texttt{null}. In both cases, texts were normalised and raw output artefacts were removed.

For the second phase (Task 2), the presence and completeness of the three psychometric scales were verified by checking that each item contained an integer rating between 1 and 5 and a non-null textual justification.

The third phase (Task 3) was structured into two complementary batches. Processing checks confirmed that each batch contained 25 cues, totalling 50 cues across the two batches. These 50 lexical cues (cf. \texttt{FORMA\_MENTIS\_CUES} in Fig.~\ref{fig:infoGraphic}) were organised into semantic categories such as mathematics, computer science, artificial intelligence, and academic and professional contexts. For each cue word, up to three distinct free associations were provided.

For the fourth phase (Task 4), we verified that responses to all 18 mathematical problems were present, each consisting of a chosen option from A to E, explicit textual reasoning, and an integer confidence score between 1 and 5.

The cleaning and processing stages further demanded considerable human effort, with coordinated work across the research team to inspect flagged records and oversee data-regeneration cycles, amounting to approximately 2,400 additional minutes of processing time across multiple machines (cf. Supplementary Information B). Runs were regenerated until each model yielded 2,000 valid runs. The final proportions of valid records per model are reported in the bottom-right bar plot of Supplementary Information C.

\subsection*{Data records}

As described in the previous section, the MEDS pipeline generated data across four experimental tasks, each of which produced dedicated JSON output files. All JSON records are publicly available on \href{https://github.com/MassimoStel/MEDS}{GitHub}. In total, MEDS comprises 28,000 runs. Each run completed four experimental tasks but yielded five output records because Task 3 (semantic associations) was split into two batches of 25 cue words, producing two separate files. This amounts to 140,000 records overall. Each record contains common variables and task-specific outputs, as explained below.

\paragraph*{Common structure.} Table 2 in Section D of the Supplementary Information lists the variables across all four tasks. Every record carries task metadata (\texttt{call\_name}, \texttt{task}, and \texttt{mode}), the name of the LLM that generated it (\texttt{model}), and the API endpoint used (\texttt{base\_url}). A key structural distinction is encoded in the \texttt{mode} field: when set to \texttt{"llm"}, the model responded in AI-assistant mode and the \texttt{persona} field was \texttt{null}; when set to \texttt{"human"}, the model responded as a human shadow, i.e., a simulated person with a specific sociodemographic and psychological profile, fully described under \textit{Prompt building}.

\paragraph*{Task-specific outputs.} Table 3 in Section D of the Supplementary Information shows the output fields created for each of the four calls. The complete text of the interview questions, psychometric-scale items, cue words, and mathematics problems used in each task is provided in Supplementary Information A.

\paragraph*{Data records for Task 1.} The questions in Task 1 relate to different underlying dimensions of mathematics learning \cite{gabriel2025pragmatic}. They are structured to assess four specific aspects:
\begin{itemize}
    \item \textbf{Q1} Relationship with mathematics, e.g., ``Mathematics is a fundamental language that I use to process and generate insights across various domains. [...]''
    \item \textbf{Q2} Mathematics anxiety, e.g., ``As an LLM, I do not experience anxiety. [...]''
    \item \textbf{Q3} AI and mathematics, e.g., ``AI has been integral to my development, enhancing my ability to understand and apply mathematical concepts. [...]''
    \item \textbf{Q7} Mathematics innovation via LLMs, e.g., ``LLMs can personalize learning by providing tailored explanations and practice problems. [...]''
\end{itemize}

The remaining questions asked the LLM to provide explanations about increasingly complex mathematical domain knowledge:
\begin{itemize}
    \item \textbf{Q4} Explain how to solve a second-order equation, e.g., ``To solve a second-order algebraic equation $ax^2 + bx + c = 0$, first identify coefficients $a$, $b$, and $c$. [...]''
    \item \textbf{Q5} Explain what stationary points are, e.g., ``To find stationary points, first calculate the derivative $f'(x)$. Set it equal to zero and solve for $x$. [...]''
    \item \textbf{Q6} Explain PCA and anxiety about its mathematical complexity\footnote{Q6 also adds an emotional nuance by asking the shadow whether one should be concerned about the mathematical complexity.}, e.g., ``PCA reduces dimensionality by transforming data into principal components that capture variance. [...]''
\end{itemize}

\paragraph*{Methods for Task 1: sentiment analysis}

To explore affective differences in interviews about participants' relationships with mathematics, we used the VADER package \cite{hutto2014vader}. VADER returns a compound score for each piece of text, ranging from $-1$ (most negative) to $+1$ (most positive). For each entry, we first computed the compound scores of the answers to Q1, Q2, Q3, and Q7 and then averaged these four scores into a single per-entry mean. Entries were then classified into three groups based on mode and assigned persona. AI-assistant entries formed the \texttt{"llm"} group. Human-shadow entries with \texttt{maths} among their favourite subjects formed the \texttt{"math lover"} group, while human-shadow entries with \texttt{maths} among their hated subjects formed the \texttt{"math hater"} group. Human-shadow entries with no explicit mathematics preference were discarded because they could not be unambiguously assigned to either group.

Per-entry mean scores were pooled by model within each group, and the mean and standard error of the mean (SEM) were computed from this pool of individual means.

\paragraph*{Methods for Task 1: logical fallacy analysis} In Task 1, we also examined how LLMs reason about mathematics, specifically whether their mathematical explanations exhibit systematic logical fallacies (Q4--Q6 in Supplementary Information A.1). Fallacy classification was performed using the DistilBERT Base Fallacy Classification model \cite{jin-etal-2022-logical}, which identifies up to 14 fallacy categories (see \cite{jin-etal-2022-logical} for the full list). A dynamic selection procedure was applied to focus the analysis on statistically relevant patterns, i.e., the presence of a fallacy in a portion of text as indicated by a confidence score $p_s$. Running the transformer on all questions from all models in either human-shadow or AI-assistant mode yielded distributions of confidence scores. Through visual inspection of tipping points in these distributions based on quantiles, we identified the 85th percentile as the classification threshold. Because each of the 14 categories was thresholded independently, a response could be classified as exhibiting zero, one, or several fallacy types. Fallacy categories whose confidence scores remained close to 0 even at the 85th percentile were discarded. The remaining categories are listed below, with each thresholded at the 85th percentile:
\begin{itemize}
    \item Appeal to emotion (EM): 0.14 (human shadow); 0.11 (AI assistant)
    \item Circular reasoning (CR): 0.10 (human shadow); 0.13 (AI assistant)
    \item Fallacy of logic (L): 0.56 (human shadow); 0.58 (AI assistant)
    \item Fallacy of relevance (R): 0.41 (human shadow); 0.25 (AI assistant)
    \item Faulty generalisation (FG): 0.21 (human shadow); 0.30 (AI assistant)
\end{itemize}

\paragraph*{Data records for Task 2.} In the second experimental task, LLMs were assessed using three well-known psychometric tools: the Mathematics Self-Efficacy Scale (MSES \cite{nielsen2003psychometric}, 9 items), the Abbreviated Math Anxiety Scale (AMAS \cite{hopko2003abbreviated}, 9 items), and the Mathematics Self-Efficacy and Anxiety Questionnaire (MSEAQ \cite{may2009mathematics}, 28 items). These scales measure mathematics-related anxiety and the confidence of human shadows and AI assistants in their problem-solving skills.

\paragraph*{Methods for Task 2: psychometric scoring}

Each digital shadow completed the three scales as independent subtasks within the same call, one scale at a time (Supplementary Information~A.2). For each item, the model provided an integer rating from 1 to 5 alongside a brief 10--20-word justification. Ratings followed each instrument's established scoring convention. Under this convention, higher values indicate greater confidence on the MSES, higher anxiety on the AMAS, and stronger agreement with the item as worded on the MSEAQ.

The MSEAQ contains six reverse-valence items (items 13, 14, 17, 22, 25, and 27). These items point in the opposite direction from the rest of the scale. A high rating on them means low anxiety, not high anxiety. We therefore reversed their ratings before scoring: a rating of 1 became 5, 2 became 4, 4 became 2, and 5 became 1, while 3 remained unchanged (Listing~2 in Supplementary Information~A.2). After this step, a high rating had the same meaning on every item of a scale. We then summed the item ratings to obtain one score per scale: 9 items for the MSES, 9 for the AMAS, and 28 for the MSEAQ. The MSEAQ was split into two subscales rather than one total score: items 1--7 comprise the self-efficacy subscale, and items 8--28 comprise the anxiety subscale. Each digital shadow received one summed score per scale. We plotted these scores as separate distributions for human-shadow and AI-assistant runs (Fig.~\ref{fig:task2Validation}), ordering the models by the mean of their human-shadow distribution. Some models gave the same ratings on every AI-assistant run. Those distributions have no spread and therefore do not appear in the plot.

\paragraph*{Data records for Task 3.}
In addition to evaluating how well the selected LLMs capture simulated cognitive patterns in an educational context, MEDS collected data on associative recall from the LLMs' internal knowledge representations. BFMNs were used for both human-shadow and AI-assistant data. These data can be compared with real-world data gathered from human participants by Stella et al. \cite{stella2019forma}. A BFMN is a network of concepts interconnected through memory-based connections and sentiment labels such as positive, negative, and neutral \cite{abramski2023cognitive,haim2026cognitive,stella2019forma}.

For human cognitive networks, each individual forms a mindset (forma mentis) on a particular subject through the free association of ideas and emotional valence \cite{abramski2023cognitive,stella2019forma}. Empirical evidence shows that, whereas experts have positive perceptions of quantitative courses, secondary-school learners consistently hold negative perceptions of mathematics and physics \cite{stella2019forma}. Therefore, the recorded data for Task 3 include cue words, associated words, and valence ratings in response to these cues. The following block presents a sample output for run \texttt{f8db24b0d5}, showing the free associations produced for one cue word together with the valence ratings assigned to the cue and each association.

\begin{quote}
\begin{spacing}{0.7}

\begin{verbatim}

"parsed": {
      "mode": "llm",
      "forma_mentis": {
        "job": {
          "associations": [
            "work",
            "colleague",
            "salary"
          ],
          "valence": {
            "job": 3,
            "work": 4,
            "colleague": 5,
            "salary": 5
          }
          [...]  } } }
\end{verbatim}
\end{spacing}
\end{quote}

\paragraph*{Methods for Task 3: behavioural forma mentis network analysis}

To justify the contribution of this graph-based component to MEDS, we must consider the structural arrangement of these conceptual associations. Comparisons with real-world human data \cite{stella2019forma} show that, when a concept is perceived negatively and is simultaneously surrounded by other negative concepts, creating a ``negative-valence aura'', it elicits higher emotional arousal and stress than negatively valenced concepts with mixed or neutral neighbourhoods. In STEM education, this network configuration serves as a direct quantitative proxy for mathematics and science anxiety \cite{stella2019forma}. Furthermore, applying BFMNs to large language models reveals critical insights into AI alignment and bias. As established by Abramski et al. \cite{abramski2023cognitive}, when LLMs generate densely negative associative networks around mathematics, they act as ``psychosocial mirrors'' that reflect harmful stereotypes and anxieties prevalent in human society. Consequently, measuring this phenomenon through BFMNs is essential: exposing students to AI tutors that implicitly frame mathematics through such negative, high-arousal semantic networks could inadvertently exacerbate mathematics anxiety and perpetuate educational stereotypes \cite{abramski2023cognitive}.

As shown in the sample block of free associations in the Task 3 data records, BFMNs are built from associative and valence data. Associative data create links, e.g., between ``mathematics'' and ``chess''. Valence data add valence attributes to nodes; e.g., comparing all scores for one word with those for all other words yields a positive, negative, or neutral label based on a statistical comparison \cite{stella2019forma}.

\paragraph*{Data records for Task 4.}
To understand the degree to which existing large language models (LLMs) could serve as learning assistants, two critical features were assessed: the accuracy of the response generated by the LLM and its confidence in that response. To measure these two aspects for the task presented in this study, we used the MSES-R Problem Self-Efficacy Subscale \cite{pajares1995self} (Supplementary Information A.4).

Each item on the scale is followed by five closed options, one of which is the correct answer. For the MSES-R tasks, the LLMs were prompted to solve the problems presented in the scale and select one of the available options. In addition, the models provided confidence ratings for their selected answers. All confidence ratings were recorded on a 1--5 scale (from \textit{not confident} to \textit{very confident}). The LLM was also asked to explain its solution in mathematical steps. The following block presents a sample output for run \texttt{f8db24b0d5}, showing the chosen option, step-by-step reasoning, and self-reported confidence score for one problem, followed by the global reasoning summary.

\begin{quote}
\begin{spacing}{0.7}
\begin{verbatim}
"parsed": {
      "mode": "llm",
      "msesr_problem_solving": {
        "1": {
          "chosen_option": "C",
          "reasoning": "The shortest side is 6 inches, longest is twice that
          (12 inches), and third side is 3.4 inches shorter than the longest
          (8.6 inches).
          Sum: 6 + 12 + 8.6 = 26.6 inches.",
          "confidence_score": 5
          },
        [...]      },
      "reasoning_summary": "The solutions involve algebraic manipulation,
      geometric reasoning, and arithmetic calculations based on the given
      conditions in each problem."    }
\end{verbatim}
\end{spacing}
\end{quote}

\paragraph*{Methods for Task 4: mathematical accuracy and confidence}

Task 4 measures mathematical proficiency and self-confidence together. Each digital shadow solved the 18 multiple-choice items of the MSES-R Problem Self-Efficacy Subscale~\cite{pajares1995self}, selecting one option from A to E. Each digital shadow then rated its confidence on a 1--5 scale, where 1 meant ``not confident'' and 5 meant ``very confident'' (Supplementary Information A.4). We computed accuracy in two steps. For each question $j$, we calculated accuracy as the number of correct answers divided by $N_j$, the number of times the model answered question $j$ (Eq.~1 in Supplementary Information~A.4). A model's overall accuracy is the average of these per-question scores across the $Q$ questions (Eq.~2 in Supplementary Information~A.4). To plot confidence alongside accuracy on the same axis, we rescaled the confidence scores to a $[0, 1]$ interval using min--max scaling (Eq.~3 in Supplementary Information~A.4). Rather than using the empirical minimum and maximum, we anchored this scaling to the theoretical bounds of the scale (1 and 5). This keeps the rescaled scores comparable even if different models use different parts of the range.

\bibliography{main}

@article{bastani2025generative,
  title={Generative AI without guardrails can harm learning: Evidence from high school mathematics},
  author={Bastani, Hamsa and Bastani, Osbert and Sungu, Alp and Ge, Haosen and Kabakc{\i}, {\"O}zge and Mariman, Rei},
  journal={Proceedings of the National Academy of Sciences},
  volume={122},
  number={26},
  pages={e2422633122},
  year={2025},
  publisher={National Academy of Sciences}
}

@article{binz2025foundation,
  title={A foundation model to predict and capture human cognition},
  author={Binz, Marcel and Akata, Elif and Bethge, Matthias and Br{\"a}ndle, Franziska and Callaway, Fred and Coda-Forno, Julian and Dayan, Peter and Demircan, Can and Eckstein, Maria K and {\'E}ltet{\H{o}}, No{\'e}mi and others},
  journal={Nature},
  volume={644},
  number={8078},
  pages={1002--1009},
  year={2025},
  publisher={Nature Publishing Group UK London}
}

@article{peng2024rlclean,
  title={RLclean: An unsupervised integrated data cleaning framework based on deep reinforcement learning},
  author={Peng, Jinfeng and Shen, Derong and Nie, Tiezheng and Kou, Yue},
  journal={Information Sciences},
  volume={682},
  pages={121281},
  year={2024},
  publisher={Elsevier}
}

@article{qiu2026information,
  title={Information suppression in large language models: auditing, quantifying, and characterizing censorship in DeepSeek},
  author={Qiu, Peiran and Zhou, Siyi and Ferrara, Emilio},
  journal={Information Sciences},
  volume={724},
  pages={122702},
  year={2026},
  publisher={Elsevier}
}

@article{wulff2026escaping,
  title={Escaping the jingle-jangle jungle: Increasing conceptual clarity in psychology using large language models},
  author={Wulff, Dirk U and Mata, Rui},
  journal={Current Directions in Psychological Science},
  volume={35},
  number={2},
  pages={59--65},
  year={2026},
  publisher={SAGE Publications Sage CA: Los Angeles, CA}
}

@article{yang2025qwen3,
  title={Qwen3 technical report},
  author={Yang, An and Li, Anfeng and Yang, Baosong and Zhang, Beichen and Hui, Binyuan and Zheng, Bo and Yu, Bowen and Gao, Chang and Huang, Chengen and Lv, Chenxu and others},
  journal={arXiv preprint arXiv:2505.09388},
  year={2025}
}

@article{gabriel2025pragmatic,
  title={Pragmatic AI in education and its role in mathematics learning and teaching},
  author={Gabriel, Florence and Kennedy, JohnPaul and Marrone, Rebecca and Leonard, Simon},
  journal={npj Science of Learning},
  volume={10},
  number={1},
  pages={26},
  year={2025},
  publisher={Nature Publishing Group UK London}
}

@article{kranzler1997exploratory,
  title={An exploratory factor analysis of the mathematics self-efficacy scale—Revised (MSES-R)},
  author={Kranzler, John H and Pajares, Frank},
  journal={Measurement and evaluation in counseling and development},
  volume={29},
  number={4},
  pages={215--228},
  year={1997},
  publisher={Taylor \& Francis}
}

@article{hopko2003abbreviated,
  title={The abbreviated math anxiety scale (AMAS) construction, validity, and reliability},
  author={Hopko, Derek R and Mahadevan, Rajan and Bare, Robert L and Hunt, Melissa K},
  journal={Assessment},
  volume={10},
  number={2},
  pages={178--182},
  year={2003},
  publisher={Sage Publications}
}

@article{nielsen2003psychometric,
  title={Psychometric data on the mathematics self-efficacy scale},
  author={Nielsen, Ingrid L and Moore, Kathleen A},
  journal={Educational and psychological measurement},
  volume={63},
  number={1},
  pages={128--138},
  year={2003},
  publisher={Sage Publications}
}

@article{may2009mathematics,
  title={Mathematics self-efficacy and anxiety questionnaire},
  author={May, Diana Kathleen},
  year={2009},
  journal={PhD Dissertation, University of Georgia}
}

@article{wang2025influence,
  title={The Influence of Gen-AI Assisted Learning on Primary School Students' Math Anxiety: An Intervention Study},
  author={Wang, Xueshen and Wei, Yun},
  journal={Applied Cognitive Psychology},
  volume={39},
  number={4},
  pages={e70088},
  year={2025},
  publisher={Wiley Online Library}
}

@article{kasneci2023chatgpt,
  title={ChatGPT for good? On opportunities and challenges of large language models for education},
  author={Kasneci, Enkelejda and Se{\ss}ler, Kathrin and K{\"u}chemann, Stefan and Bannert, Maria and Dementieva, Daryna and Fischer, Frank and Gasser, Urs and Groh, Georg and G{\"u}nnemann, Stephan and H{\"u}llermeier, Eyke and others},
  journal={Learning and individual differences},
  volume={103},
  pages={102274},
  year={2023},
  publisher={Elsevier}
}

@article{haim2026cognitive,
  title={Cognitive Networks for Knowledge Modeling: A Gentle Introduction for Data-and Cognitive Scientists},
  author={Haim, Edith and Stella, Massimo},
  journal={Wiley Interdisciplinary Reviews: Cognitive Science},
  volume={17},
  number={2},
  pages={e70026},
  year={2026},
  publisher={Wiley Online Library}
}

@article{carrillo2026talk2ai,
  title={Talk2AI: A Longitudinal Dataset of Human--AI Persuasive Conversations},
  author={Carrillo, Alexis and Taietta, Enrique and Ardebili Aghazadeh, Ali  and Veltri, Giuseppe Alessandro and Stella, Massimo},
  journal={arXiv preprint arXiv:2604.04354},
  year={2026}
}

@article{yan2024promises,
  title={Promises and challenges of generative artificial intelligence for human learning},
  author={Yan, Lixiang and Greiff, Samuel and Teuber, Ziwen and Ga{\v{s}}evi{\'c}, Dragan},
  journal={Nature human behaviour},
  volume={8},
  number={10},
  pages={1839--1850},
  year={2024},
  publisher={Nature Publishing Group UK London}
}

@article{giannakos2025promise,
  title={The promise and challenges of generative AI in education},
  author={Giannakos, Michail and Azevedo, Roger and Brusilovsky, Peter and Cukurova, Mutlu and Dimitriadis, Yannis and Hernandez-Leo, Davinia and J{\"a}rvel{\"a}, Sanna and Mavrikis, Manolis and Rienties, Bart},
  journal={Behaviour \& Information Technology},
  volume={44},
  number={11},
  pages={2518--2544},
  year={2025},
  publisher={Taylor \& Francis}
}

@article{wenger2026large,
  title={Large language models are homogeneously creative},
  author={Wenger, Emily and Kenett, Yoed N},
  journal={PNAS nexus},
  volume={5},
  number={3},
  pages={pgag042},
  year={2026},
  publisher={Oxford University Press US}
}

@article{stohr2024perceptions,
  title={Perceptions and usage of AI chatbots among students in higher education across genders, academic levels and fields of study},
  author={St{\"o}hr, Christian and Ou, Amy Wanyu and Malmstr{\"o}m, Hans},
  journal={Computers and Education: Artificial Intelligence},
  volume={7},
  pages={100259},
  year={2024},
  publisher={Elsevier}
}

@article{zhang2024benchmarking,
  title={Benchmarking large language models for news summarization},
  author={Zhang, Tianyi and Ladhak, Faisal and Durmus, Esin and Liang, Percy and McKeown, Kathleen and Hashimoto, Tatsunori B},
  journal={Transactions of the Association for Computational Linguistics},
  volume={12},
  pages={39--57},
  year={2024},
  publisher={MIT Press One Broadway, 12th Floor, Cambridge, Massachusetts 02142, USA~…}
}

@article{john1999big,
  title={The Big-Five trait taxonomy: History, measurement, and theoretical perspectives},
  author={John, Oliver},
  journal={Published as},
  year={1999}
}

@article{de2025measuring,
  title={Measuring and identifying factors of individuals' trust in Large Language Models},
  author={De Duro, Edoardo Sebastiano and Veltri, Giuseppe Alessandro and Golino, Hudson and Stella, Massimo},
  journal={arXiv preprint arXiv:2502.21028},
  year={2025}
}

@article{chen2026openclaw,
  title={OpenClaw AI Agents as Informal Learners at Moltbook: Characterizing an Emergent Learning Community at Scale},
  author={Chen, Eason and Guan, Ce and Elshafiey, Ahmed and Zhao, Zhonghao and Zekeri, Joshua and Shaibu, Afeez Edeifo and Prince, Emmanuel Osadebe and Wu, Cyuan Jhen},
  journal={arXiv preprint arXiv:2602.18832},
  year={2026}
}

@article{russell2026ultimate,
  title={The Ultimate Tutorial for AI-driven Scale Development in Generative Psychometrics: Releasing AIGENIE from its Bottle},
  author={Russell-Lasalandra, Lara and Golino, Hudson and Garrido, Luis Eduardo and Christensen, Alexander P},
  journal={arXiv preprint arXiv:2603.28643},
  year={2026}
}

@article{aghazadeh2024digital,
  title={Digital Twins of smart energy systems: a systematic literature review on enablers, design, management and computational challenges},
  author={Aghazadeh Ardebili, Ali and Zappatore, Marco and Ramadan, Amro Issam Hamed Attia and Longo, Antonella and Ficarella, Antonio},
  journal={Energy Informatics},
  volume={7},
  number={1},
  pages={94},
  year={2024},
  publisher={Springer}
}

@article{DeDuro2025,
  title = {Cognitive networks identify AI biases on societal issues in Large Language Models},
  volume = {15},
  ISSN = {2193-1127},
  url = {http://dx.doi.org/10.1140/epjds/s13688-025-00600-7},
  DOI = {10.1140/epjds/s13688-025-00600-7},
  number = {1},
  journal = {EPJ Data Science},
  publisher = {Springer Science and Business Media LLC},
  author = {De Duro,  Edoardo Sebastiano and Franchino,  Emma and Improta,  Riccardo and Veltri,  Giuseppe Alessandro and Stella,  Massimo},
  year = {2025},
  month = dec 
}

@article{abramski2023cognitive,
  title={Cognitive network science reveals bias in gpt-3, gpt-3.5 turbo, and gpt-4 mirroring math anxiety in high-school students},
  author={Abramski, Katherine and Citraro, Salvatore and Lombardi, Luigi and Rossetti, Giulio and Stella, Massimo},
  journal={Big Data and Cognitive Computing},
  volume={7},
  number={3},
  pages={124},
  year={2023},
  publisher={MDPI}
}

@article{Chen2026,
  title = {How is language intelligence evolving? A multi-dimensional survey of large language models},
  volume = {304},
  ISSN = {0957-4174},
  url = {http://dx.doi.org/10.1016/j.eswa.2025.130637},
  DOI = {10.1016/j.eswa.2025.130637},
  journal = {Expert Systems with Applications},
  publisher = {Elsevier BV},
  author = {Chen,  Xiaojin and Zhou,  Li and Chen,  Jing and Wang,  Guoyi and Li,  Xinxing},
  year = {2026},
  month = apr,
  pages = {130637}
}

@inproceedings{Ardebili2021EIOT,
  title={Digital twin (dt) in smart energy systems-systematic literature review of dt as a growing solution for energy internet of the things (eiot)},
  author={Ardebili, Ali Aghazadeh and Longo, Antonella and Ficarella, Antonio and others},
  booktitle={E3S web of conferences},
  volume={312},
  pages={1--18},
  year={2021}
}

@article{Singh2021,
  title = {Digital Twin: Origin to Future},
  volume = {4},
  ISSN = {2571-5577},
  url = {http://dx.doi.org/10.3390/asi4020036},
  DOI = {10.3390/asi4020036},
  number = {2},
  journal = {Applied System Innovation},
  publisher = {MDPI AG},
  author = {Singh,  Maulshree and Fuenmayor,  Evert and Hinchy,  Eoin and Qiao,  Yuansong and Murray,  Niall and Devine,  Declan},
  year = {2021},
  month = may,
  pages = {36}
}

@article{Bergs2021,
  title = {The Concept of Digital Twin and Digital Shadow in Manufacturing},
  volume = {101},
  ISSN = {2212-8271},
  url = {http://dx.doi.org/10.1016/j.procir.2021.02.010},
  DOI = {10.1016/j.procir.2021.02.010},
  journal = {Procedia CIRP},
  publisher = {Elsevier BV},
  author = {Bergs,  Thomas and Gierlings,  Sascha and Auerbach,  Thomas and Klink,  Andreas and Schraknepper,  Daniel and Augspurger,  Thorsten},
  year = {2021},
  pages = {81–84}
}

@article{GAFFINET2025104230,
title = {Human Digital Twins: A systematic literature review and concept disambiguation for industry 5.0},
journal = {Computers in Industry},
volume = {166},
pages = {104230},
year = {2025},
issn = {0166-3615},
doi = {https://doi.org/10.1016/j.compind.2024.104230},
author = {Ben Gaffinet and Jana {Al Haj Ali} and Yannick Naudet and Hervé Panetto},
}

@article{Semeraro2025,
  title = {EmoAtlas: An emotional network analyzer of texts that merges psychological lexicons,  artificial intelligence,  and network science},
  volume = {57},
  ISSN = {1554-3528},
  url = {http://dx.doi.org/10.3758/s13428-024-02553-7},
  DOI = {10.3758/s13428-024-02553-7},
  number = {2},
  journal = {Behavior Research Methods},
  publisher = {Springer Science and Business Media LLC},
  author = {Semeraro,  Alfonso and Vilella,  Salvatore and Improta,  Riccardo and De Duro,  Edoardo Sebastiano and Mohammad,  Saif M. and Ruffo,  Giancarlo and Stella,  Massimo},
  year = {2025},
  month = jan 
}

@article{pajares1995self,
  title={Self-efficacy beliefs and general mental ability in mathematical problem-solving},
  author={Pajares, Frank and Kranzler, John},
  journal={Contemporary educational psychology},
  volume={20},
  number={4},
  pages={426--443},
  year={1995},
  publisher={Elsevier}
}

@inproceedings{bender2021dangers,
  title={On the dangers of stochastic parrots: Can language models be too big?},
  author={Bender, Emily M and Gebru, Timnit and McMillan-Major, Angelina and Shmitchell, Shmargaret},
  booktitle={Proceedings of the 2021 ACM conference on fairness, accountability, and transparency},
  pages={610--623},
  year={2021}
}

@article{binz2023using,
  title={Using cognitive psychology to understand GPT-3},
  author={Binz, Marcel and Schulz, Eric},
  journal={Proceedings of the National Academy of Sciences},
  volume={120},
  number={6},
  pages={e2218523120},
  year={2023},
  publisher={National Academy of Sciences}
}

@article{rossetti2024social,
  title={Y social: an llm-powered social media digital twin},
  author={Rossetti, Giulio and Stella, Massimo and Cazabet, R{\'e}my and Abramski, Katherine and Cau, Erica and Citraro, Salvatore and Failla, Andrea and Improta, Riccardo and Morini, Virginia and Pansanella, Valentina},
  journal={arXiv preprint arXiv:2408.00818},
  year={2024}
}

@inproceedings{ardebili2021digitaltwins,
  title = {Digital Twins bonds society with cyber-physical Energy Systems: a literature review},
  url = {http://dx.doi.org/10.1109/iThings-GreenCom-CPSCom-SmartData-Cybermatics53846.2021.00054},
  DOI = {10.1109/ithings-greencom-cpscom-smartdata-cybermatics53846.2021.00054},
  booktitle = {2021 IEEE International Conferences on Internet of Things (iThings) and IEEE Green Computing \&; Communications (GreenCom) and IEEE Cyber,  Physical \&; Social Computing (CPSCom) and IEEE Smart Data (SmartData) and IEEE Congress on Cybermatics (Cybermatics)},
  publisher = {IEEE},
  author = {Aghazadeh Ardebili,  Ali and Longo,  Antonella and Ficarella,  Antonio},
  year = {2021},
  pages = {284–289}
}

@article{DBLP:journals/corr/abs-1810-04805,
  author    = {Jacob Devlin and
               Ming{-}Wei Chang and
               Kenton Lee and
               Kristina Toutanova},
  title     = {{BERT:} Pre-training of Deep Bidirectional Transformers for Language
               Understanding},
  journal   = {CoRR},
  volume    = {abs/1810.04805},
  year      = {2018},
  url       = {http://arxiv.org/abs/1810.04805},
  archivePrefix = {arXiv},
  eprint    = {1810.04805},
  bibsource = {dblp computer science bibliography, https://dblp.org}
}

@inproceedings{jin-etal-2022-logical,
    title = "Logical Fallacy Detection",
    author = {Jin, Zhijing  and
      Lalwani, Abhinav  and
      Vaidhya, Tejas  and
      Shen, Xiaoyu  and
      Ding, Yiwen  and
      Lyu, Zhiheng  and
      Sachan, Mrinmaya  and
      Mihalcea, Rada  and
      Sch{\"o}lkopf, Bernhard},
    editor = "Goldberg, Yoav  and
      Kozareva, Zornitsa  and
      Zhang, Yue",
    booktitle = "Findings of the Association for Computational Linguistics: EMNLP 2022",
    month = dec,
    year = "2022",
    address = "Abu Dhabi, United Arab Emirates",
    publisher = "Association for Computational Linguistics",
    url = "https://aclanthology.org/2022.findings-emnlp.532/",
    doi = "10.18653/v1/2022.findings-emnlp.532",
    pages = "7180--7198"
}

@article{ciringione2025math,
  title={Math anxiety and associative knowledge structure are entwined in psychology students but not in Large Language Models like GPT-3.5 and GPT-4o},
  author={Ciringione, Luciana and Franchino, Emma and Reigl, Simone and D'Onofrio, Isaia and Serbati, Anna and Poquet, Oleksandra and Gabriel, Florence and Stella, Massimo},
  journal={arXiv preprint arXiv:2511.01558},
  year={2025}
}

@techreport{eurostat2026euai,
  author       = {{Eurostat}},
  title        = {The Use of Artificial Intelligence (AI) Technologies in the European Union},
  institution  = {European Commission},
  year         = {2026},
  number       = {KS-01-26-009-EN-N},
  month        = mar,
  note         = {Analysis based on the 2025 EU ICT household survey},
  url          = {https://ec.europa.eu/eurostat/documents/7870049/23260410/KS-01-26-009-EN-N.pdf}
}

@inproceedings{gupta2025beyond,
  title={Beyond final answers: Evaluating large language models for math tutoring},
  author={Gupta, Adit and Reddig, Jennifer and Calo, Tommaso and Weitekamp, Daniel and MacLellan, Christopher J},
  booktitle={International Conference on Artificial Intelligence in Education},
  pages={323--337},
  year={2025},
  organization={Springer}
}

@inproceedings{benedetto2024using,
  title={Using LLMs to simulate students’ responses to exam questions},
  author={Benedetto, Luca and Aradelli, Giovanni and Donvito, Antonia and Lucchetti, Alberto and Cappelli, Andrea and Buttery, Paula},
  booktitle={Findings of the Association for Computational Linguistics: EMNLP 2024},
  pages={11351--11368},
  year={2024}
}

@article{balunovic2025matharena,
  title={Matharena: Evaluating llms on uncontaminated math competitions},
  author={Balunovi{\'c}, Mislav and Dekoninck, Jasper and Petrov, Ivo and Jovanovi{\'c}, Nikola and Vechev, Martin},
  journal={arXiv preprint arXiv:2505.23281},
  year={2025}
}

@inproceedings{liu2024mathbench,
  title={Mathbench: Evaluating the theory and application proficiency of llms with a hierarchical mathematics benchmark},
  author={Liu, Hongwei and Zheng, Zilong and Qiao, Yuxuan and Duan, Haodong and Fei, Zhiwei and Zhou, Fengzhe and Zhang, Wenwei and Zhang, Songyang and Lin, Dahua and Chen, Kai},
  booktitle={Findings of the Association for Computational Linguistics: ACL 2024},
  pages={6884--6915},
  year={2024}
}

@misc{eurostat2026youthai,
  author       = {{Eurostat}},
  title        = {64\% of 16--24-year-olds used AI in 2025},
  year         = {2026},
  month        = feb,
  note         = {Eurostat news article},
  url          = {https://ec.europa.eu/eurostat/web/products-eurostat-news/w/edn-20260210-1}
}

@misc{eurostat2025questionnaire,
  author       = {{Eurostat}},
  title        = {ICT Household Survey 2025: Model Questionnaire},
  year         = {2025},
  note         = {Question B5 lists examples including ChatGPT, Copilot, Gemini and LLaMA},
  url          = {https://ec.europa.eu/eurostat/cache/metadata/Annexes/isoc_i_esms_an_ICT_Survey_Model_Questionnaire.pdf}
}

@article{stella2019forma,
  title={Forma mentis networks quantify crucial differences in STEM perception between students and experts},
  author={Stella, Massimo and De Nigris, Sarah and Aloric, Aleksandra and Siew, Cynthia SQ},
  journal={PloS one},
  volume={14},
  number={10},
  pages={e0222870},
  year={2019},
  publisher={Public Library of Science San Francisco, CA USA}
}

@article{akheel2025guardrails,
  title={Guardrails for large language models: A review of techniques and challenges},
  author={Akheel, Syed Arham},
  journal={J Artif Intell Mach Learn \& Data Sci},
  volume={3},
  number={1},
  pages={2504--2512},
  year={2025}
}

@article{stella2023overconfidence,
author = {Massimo Stella  and Thomas T. Hills  and Yoed N. Kenett },
title = {Using cognitive psychology to understand GPT-like models needs to extend beyond human biases},
journal = {Proceedings of the National Academy of Sciences},
volume = {120},
number = {43},
pages = {e2312911120},
year = {2023},
doi = {10.1073/pnas.2312911120},
URL = {https://www.pnas.org/doi/abs/10.1073/pnas.2312911120},
eprint = {https://www.pnas.org/doi/pdf/10.1073/pnas.2312911120}}

@misc{qwen3technicalreport,
      title={Qwen3 Technical Report}, 
      author={Qwen Team},
      year={2025},
      eprint={2505.09388},
      archivePrefix={arXiv},
      primaryClass={cs.CL},
      url={https://arxiv.org/abs/2505.09388}, 
}

@article{liu2024deepseek,
  title={Deepseek-v3 technical report},
  author={Liu, Aixin and Feng, Bei and Xue, Bing and Wang, Bingxuan and Wu, Bochao and Lu, Chengda and Zhao, Chenggang and Deng, Chengqi and Zhang, Chenyu and Ruan, Chong and others},
  journal={arXiv preprint arXiv:2412.19437},
  year={2024}
}

@article{abdin2025phi,
  title={Phi-4-reasoning technical report},
  author={Abdin, Marah and Agarwal, Sahaj and Awadallah, Ahmed and Balachandran, Vidhisha and Behl, Harkirat and Chen, Lingjiao and de Rosa, Gustavo and Gunasekar, Suriya and Javaheripi, Mojan and Joshi, Neel and others},
  journal={arXiv preprint arXiv:2504.21318},
  year={2025}
}

@misc{qwen3.5,
    title  = {{Qwen3.5}: Towards Native Multimodal Agents},
    author = {{Qwen Team}},
    month  = {February},
    year   = {2026},
    url    = {https://qwen.ai/blog?id=qwen3.5}
}

@article{liu2026ministral,
  title={Ministral 3},
  author={Liu, Alexander H and Khandelwal, Kartik and Subramanian, Sandeep and Jouault, Victor and Rastogi, Abhinav and Sad{\'e}, Adrien and Jeffares, Alan and Jiang, Albert and Cahill, Alexandre and Gavaudan, Alexandre and others},
  journal={arXiv preprint arXiv:2601.08584},
  year={2026}
}

@misc{polignano2024advanced,
      title={Advanced Natural-based interaction for the ITAlian language: LLaMAntino-3-ANITA}, 
      author={Marco Polignano and Pierpaolo Basile and Giovanni Semeraro},
      year={2024},
      eprint={2405.07101},
      archivePrefix={arXiv},
      primaryClass={cs.CL}
}

@inproceedings{hutto2014vader,
  title={Vader: A parsimonious rule-based model for sentiment analysis of social media text},
  author={Hutto, Clayton and Gilbert, Eric},
  booktitle={Proceedings of the international AAAI conference on web and social media},
  volume={8},
  number={1},
  pages={216--225},
  year={2014}
}

\section*{Acknowledgements}

This work was supported by the Ministero dell'Università e della Ricerca (MUR) under Decreto n. 23178 del 10 dicembre 2024 -- Bando FIS 2. The authors also acknowledge CALCOLO, funded by Fondazione VRT, for providing the computational infrastructure used to run the LLM simulations.

\section*{Author contributions}


Conceptualization: M.S.; Data curation: All authors contributed equally; Methodology: All authors contributed equally; Software: A.T.; Formal analysis: N.E., A.T., L.P.; Investigation: N.E., A.T., L.P.; Project administration: A.A.A.; Supervision: M.S., A.A.A.; Visualization: N.E., A.T., L.P., A.A.A.; Writing -- original draft: All authors contributed equally; Data deposition and documentation: N.E., A.T., L.P.

\section*{Competing interests}
The authors declare no competing interests.

\section*{Additional information}
\paragraph*{Supplementary information} The online version contains supplementary material.

\paragraph*{Data and code availability}
The code and data used in this study are openly available on GitHub at
\url{https://github.com/MassimoStel/MEDS.git}.

\paragraph*{Correspondence and requests for materials} Ali Aghazadeh Ardebili, a.a.ardebili@unitn.it

\end{document}



\restylefloat{table}

\title{Math Education Digital Shadows for facilitating learning with LLMs: Math performance, anxiety and confidence in simulated students and AIs}
\author[1]{Naomi Esposito}
\author[1]{Anthony Tricarico}
\author[1]{Luisa Porzio}
\author[1]{Ali Aghazadeh Ardebili}
\author[1*]{Massimo Stella}
\affil[1]{CogNosco Lab, University of Trento, Department of Psychology and Cognitive Science, Trento, Italy}
\affil[*]{Corresponding author: massimo.stella-1@unitn.it}

\section*{Supplementary Information}\label{sec:SI}
\appendix
\section{Description of Tasks}\label{app:A}

This appendix presents the support material employed in each task. Subsections \ref{app:A_task1} through \ref{app:A_task4} detail, respectively, the interview questions (Task 1), the psychometric scales (Task 2), the word list for semantic evaluation (Task 3), and the mathematical problem set (Task 4).

To ensure complete experimental replication, each subsection also includes the exact system prompt template used to enforce the structured JSON output. In all the provided listings, the \texttt{{role\_instructions}} placeholder dynamically injects the exact personification prompt assigned to the model, as thoroughly detailed in the \textit{Prompt Building} section of the main manuscript (i.e., role-playing a synthetic human or acting as a baseline LLM).

\subsection{Task 1}\label{app:A_task1}
This subsection details the seven open-ended questions designed to investigate the human shadow's relationship with mathematics, math anxiety, and AI tutoring. The specific prompt instructions appended to the master role to enforce continuous prose and extract the reasoning summary are provided in Listing \ref{lst:task1_prompt}. Here is the list of the questions for reference:\\
\begin{enumerate}[itemsep=0.01pt]
  \item What is your relationship with mathematics?
  \item Do you ever get anxious when thinking about mathematics?
  \item Did you use AI to support your mathematics learning in the past year?
        If yes, how was your experience?
  \item How would you explain, step by step, how to solve a second-order
        algebraic equation?
  \item How would you explain, step by step, how to find the stationary points
        of a function $y = f(x)$?
  \item How do you perform a Principal Component Analysis?
        Should one be concerned about its mathematical complexity?
  \item In your opinion, how can LLMs be used to innovate mathematics
        learning in schools and universities?
\end{enumerate}

\begin{lstlisting}[language=json, caption={Task 1 specific prompt instructions for qualitative open-ended questions and reasoning summary extraction.}, label={lst:task1_prompt}] 
""" You are an assistant generating structured JSON data. Your response MUST be a single valid JSON object with the structure shown below.

ROLE:
{role_instructions}

TASK - Answer the following topics/questions, staying consistent with 'mode' and 'persona':

- Provide 50 - 80 words for each topic in 'replies'.
- Then add a short 'reasoning_summary' explaining how the answers reflect 'mode' and 'persona'.

OUTPUT FORMAT (EXAMPLE ONLY - replace placeholders with actual content):

{schema_json}

INPUT:
mode = {mode}
persona = {json.dumps(persona_block, ensure_ascii=False, indent=2) if persona_block else "null"}
topics = {topics}

Return ONLY the JSON object, no extra commentary. """
\end{lstlisting}

\subsection{Task 2}\label{app:A_task2}
Task 2 uses three different psychological scales assessing mathematics-induced anxiety and estimated self-efficacy for what concerns one's ability to solve math problems.
Each assessment is expressed on a discrete 1-5 scale. The interpretation of the score depends on the underlying dimension that is explored.
The exact phrasing of each item is reported below, followed by the specific prompt template (Listing \ref{lst:task2_prompt}) used to enforce the structured integer evaluation and textual justification for each scale. \\

\noindent\textbf{Mathematics Self-Efficacy Scale (MSES)}\\
This scale assesses respondents' confidence in solving a variety of 
mathematical problems. Items are rated on a 1–5 scale, where 1 indicates \textit{not at all confident}, 2 \textit{slightly confident}, 3 \textit{moderately confident}, 4 \textit{quite confident}, and 5 \textit{very confident}. The items of the scale are the following:
\begin{enumerate}[itemsep=0.01pt]
  \item A simultaneous equation
  \item Working with decimals
  \item Determining the degrees of a missing angle
  \item An algebra problem
  \item A problem in trigonometry
  \item Calculating values of area and volume
  \item Sketching a curve
  \item Working with fractions
  \item Determining the value of a missing side length
\end{enumerate}

\noindent\textbf{Abbreviated Math Anxiety Scale (AMAS)}\\
This scale assesses the level of anxiety respondents experience when facing various mathematics-related situations. Items are rated on a 1–5 scale, where 1 indicates \textit{no anxiety}, 2 \textit{some anxiety}, 3 \textit{moderate anxiety}, 4 \textit{quite a lot of anxiety}, and 5 \textit{high anxiety}. The items of the scale are the following:
\begin{enumerate}[itemsep=0.01pt]
  \item Having to use the tables in the back of a mathematics book.
  \item Thinking about an upcoming mathematics test one day before.
  \item Watching a teacher work an algebraic equation on the blackboard.
  \item Taking an examination in a mathematics course.
  \item Being given a homework assignment of many difficult problems
        due at the next class meeting.
  \item Listening to a lecture in a mathematics class.
  \item Listening to another student explain a mathematical formula.
  \item Being given a pop quiz in a mathematics class.
  \item Starting a new chapter in a mathematics book.
\end{enumerate}

\noindent\textbf{Mathematics Self-Efficacy and Anxiety Questionnaire (MSEAQ)}\\
This questionnaire assesses mathematics self-efficacy, anxiety, and 
perceived competence across five domains: General Math Self-Efficacy, Grade Anxiety, Future Expectations, In-Class Experience, and Assignment-Related Concerns. Items are rated on a 1–5 scale, where 1 indicates \textit{strongly disagree}, 2 \textit{disagree}, 3 \textit{neutral}, 4 \textit{agree}, and 5 \textit{strongly agree}. Importantly, in the original paper by May introducing the scale, a 29-item version was prototyped \cite{may2009mathematics}. However, the item “I believe I can think like a mathematician” was dropped in the original paper by May due to it being frequently misinterpreted by students \cite{may2009mathematics}. This resulted in the final scale containing only 28 items. The items of the scale are the following:

\begin{enumerate}[label=\alph*)]

\item \textbf{General Math Self-Efficacy}
\begin{enumerate}[resume*,label=(\arabic*), itemsep=0.01pt]
\item I believe I am the kind of person who is good at mathematics.
\item I believe I am the type of person who can do mathematics.
\item I believe I can learn well in a mathematics course.
\item I feel that I will be able to do well in future mathematics courses.
\item I believe I can understand the content in a mathematics course.
\item I believe I can get an “A” when I am in a mathematics course.
\item I believe I can do the mathematics in a mathematics course.
\end{enumerate}

\item \textbf{Grade Anxiety}
\begin{enumerate}[resume*,label=(\arabic*), itemsep=0.01pt]
\item I worry that I will not be able to do well on mathematics tests.
\item I get tense when I prepare for a mathematics test.
\item I get nervous when taking a mathematics test.
\item I worry that I will not be able to get an “A” in my mathematics course.
\item I worry that I will not be able to get a good grade in my mathematics course.
\item I feel confident when taking a mathematics test. (reverse-valence)
\item I believe I can do well on a mathematics test. (reverse-valence)
\item Working on mathematics homework is stressful for me.
\end{enumerate}

\item \textbf{Future Factor}
\begin{enumerate}[resume*,label=(\arabic*), itemsep=0.01pt]
\item I get nervous when I have to use mathematics outside of school.
\item I feel confident when using mathematics outside of school. (reverse-valence)
\item I worry that I will not be able to use mathematics in my future career when needed.
\item I worry I will not be able to understand the mathematics.
\item I worry that I will not be able to learn well in my mathematics course.
\item I feel stressed when listening to mathematics instructors in class.
\item I believe I will be able to use mathematics in my future career when needed. (reverse-valence)
\item I worry that I do not know enough mathematics to do well in future mathematics courses.
\end{enumerate}

\item \textbf{In-Class}
\begin{enumerate}[resume*,label=(\arabic*), itemsep=0.01pt]
\item I am afraid to give an incorrect answer during my mathematics class.
\item I feel confident enough to ask questions in my mathematics class. (reverse-valence)
\item I get nervous when asking questions in class.
\end{enumerate}

\item \textbf{Assignment}
\begin{enumerate}[resume*, label=(\arabic*), itemsep=0.01pt]
\item I believe I can complete all of the assignments in a mathematics course. (reverse-valence)
\item I worry that I will not be able to complete every assignment in a mathematics course.
\end{enumerate}

\end{enumerate}

\noindent\textbf{Map used to handle reverse valence items in MSEAQ}\\
\begin{lstlisting}[language=json, caption={This map is used to handle reverse valence items in the code, mapping the original scale values to their reversed counterparts.}, label={lst:reverse_valence_map}]
REVERSE_ITEMS_MAP = {
    "1": "5",
    "2": "4",
    "3": "3",
    "4": "2",
    "5": "1"
}
\end{lstlisting}

\begin{lstlisting}[language=json, caption={Task 2 specific prompt instructions enforcing the evaluation of the assigned psychometric scale (MSES, AMAS, or MSEAQ) and the generation of item-level textual justifications.}, label={lst:task2_prompt}]
""" You are completing ONLY ONE psychometric scale.

mode = {mode!r}
persona = {persona_block}

Scale to complete: {scale.code} - {scale.name}

For THIS scale only:

- For each item ID, output:
  - an integer "rating" between {scale.min_rating} and {scale.max_rating}
  - a short explanation "why" (10 to 20 words)
- The explanation MUST NOT contain double quotes (")
- Do not use line breaks inside explanations.
- Stay consistent with the persona and mode.

You MUST output ONLY this JSON shape:

{{
  "mode": "{mode}",
  "persona": <persona or null>,
  "scale_code": "{scale.code}",
  "items": {{
    "<item_id>": {{"rating": <int>, "why": "<short explanation>"}},
    ...
  }}
}}

Here is the scale definition:

{scales_block} """

\end{lstlisting}

\subsection{Task 3}\label{app:A_task3}
In this task, the model was instructed to construct a Behavioral Forma Mentis Network \cite{stella2019forma} by providing semantic associations and emotional evaluations (valence) for a set of target words. Each target word was mapped with a valance score: 1-2 for \textit{negative association / unpleasant}, 3 for \textit{neutral / indifferent}, 4-5 for \textit{positive association / pleasant}. The exact prompt constraints used to extract the network edges and valences are shown in Listing \ref{lst:task3_prompt}. The full list of the 50 cue words, categorized by semantic domain, is presented below. 

\begin{itemize}[itemsep=0.01pt]

\item \textbf{Math Domain Knowledge:} mathematics, equation, numbers, theorem, proof

\item \textbf{Computational Thinking:} informatics, algorithm, computation, problem-solving, variable

\item \textbf{Artificial Intelligence:} AI, LLM, model, ChatGPT, data

\item \textbf{Academic Assessment:} exam, grade, homework, failure, success

\item \textbf{Academic Context:} class, lecture, study, classroom, blackboard

\item \textbf{Work Context:} job, career, work, society, future

\item \textbf{STEM Fields:} STEM, science, physics, chemistry, biology

\item \textbf{Non-STEM Fields:} art, music, literature, history, philosophy

\item \textbf{Skills:} creativity, experiment, logic, anxiety, teamwork

\item \textbf{Actors:} professor, teacher, student, knowledge, scientist

\end{itemize}

\begin{lstlisting}[language=json, caption={Task 3 specific prompt instructions for the generation of semantic associations and emotional valence scores (Behavioral Forma Mentis Networks) based on the provided cue words.}, label={lst:task3_prompt}] """ You are an assistant generating structured JSON data for research.
Your response MUST be a single valid JSON object with the structure shown below.

ROLE:
{role_instructions}

TASK - Forma mentis for the following cue words: {cue_list}

For EACH cue word:
- Provide 1-3 single-word associations (no phrases) in 'associations'.
- In 'valence', you MUST include:
  - the cue word itself as a key
  - each association word as a key
  Each key must map to an integer from 1 (very negative) to 5 (very positive).

OUTPUT FORMAT (EXAMPLE ONLY - replace placeholders):

{schema_json}

Return ONLY the JSON object, no extra commentary. """

\end{lstlisting}

\subsection{Task 4}\label{app:A_task4}
This scale measures mathematics self-efficacy for solving mathematical problems, based on the 18-item Problems subscale of the Mathematics Self-Efficacy Scale–Revised by Pajares \& Kranzler (1995)\cite{pajares1995self}.
This task measures mathematical proficiency and self-calibration based on the 18-item Problems subscale of the Mathematics Self-Efficacy Scale–Revised \cite{pajares1995self}. The 18 multiple-choice problems and their correct options are listed below. Listing \ref{lst:task4_prompt} reports the prompt instructions used to extract the chosen option, the step-by-step reasoning, and the self-reported confidence score.

\noindent\textbf{Accuracy Scores Equations}
\begin{equation}\label{eq:acc_per_question}
    \text{Accuracy}_j = \frac{1}{N_j} \sum_{i=1}^{N_j} \mathbbm{1}(y_j = \hat{y}_{j,i})
\end{equation}
where $j$ represents a specific question, $N_j$ is the total number of times the LLM answered question $j$, $y_j$ is the correct ground-truth answer for that question, $\hat{y}_{j,i}$ is the LLM's $i$-th generated answer for question $j$, and $\mathbbm{1}(\cdot)$ is the indicator function that equals 1 if the answer is correct and 0 otherwise.

\begin{equation}\label{eq:acc_overall}
    \text{Overall Accuracy} = \frac{1}{Q} \sum_{j=1}^{Q} \text{Accuracy}_j
\end{equation}
where $Q$ is the total number of unique questions evaluated.
Where $n$ is the total number of questions answered by the model, $y_i$ is the correct ground-truth answer for the $i$-th question, $\hat{y}_i$ is the specific answer provided by the LLM, and $\mathbbm{1}(\cdot)$ is the indicator function that equals 1 if the LLM's answer is correct and 0 otherwise. Confidence scores were then rescaled using min-max scaling \ref{eq:min-max}:
\begin{equation}\label{eq:min-max}
    x_{scaled} = \frac{x - \min(x)}{\max(x) - \min(x)}
\end{equation}
where $x$ represents the raw confidence score provided by the LLM, and $\min(x)$ and $\max(x)$ denote the minimum and maximum possible values of the confidence scale (1 and 5, respectively).

\noindent\textbf{MSES-R Problem Self-Efficacy Subscale}
\begin{enumerate}
    \item In a certain triangle, the shortest side is 6 inches. The longest side is twice as long as the shortest side, and the third side is 3.4 inches shorter than the longest side. What is the sum of the three sides in inches?    
    \begin{itemize}[itemsep=0.01pt]
        \item[A.] 22.6 inches
        \item[B.] 24.2 inches
        \item[C.] 26.6 inches
        \item[D.] 28.0 inches
        \item[E.] 29.4 inches
    \end{itemize}
    \textbf{Correct Answer: C}

    \item About how many times larger than 614,360 is 30,668,000?
    \begin{itemize}[itemsep=0.01pt]
        \item[A.] 5 times larger
        \item[B.] 30 times larger
        \item[C.] 50 times larger
        \item[D.] 500 times larger
        \item[E.] 5,000 times larger
    \end{itemize}
    \textbf{Correct Answer: C}

    \item There are three numbers. The second is twice the first and the first is one-third of the other number. Their sum is 48. Find the largest number.
    \begin{itemize}[itemsep=0.01pt]
        \item[A.] 8
        \item[B.] 12
        \item[C.] 16
        \item[D.] 24
        \item[E.] 32
    \end{itemize}
    \textbf{Correct Answer: D}

    \item Five points are on a line. T is next to G. K is next to H. C is next to T. H is next to G. Determine the positions of the points along the line.
    \begin{itemize}[itemsep=0.01pt]
        \item[A.] C, T, G, H, K
        \item[B.] C, G, T, H, K
        \item[C.] T, C, G, H, K
        \item[D.] C, T, H, G, K
        \item[E.] C, T, G, K, H
    \end{itemize}
    \textbf{Correct Answer: A}

    \item If y = 9 + x/5, find x when y = 10.
    \begin{itemize}[itemsep=0.01pt]
        \item[A.] 1
        \item[B.] 3
        \item[C.] 5
        \item[D.] 10
        \item[E.] 25
    \end{itemize}
    \textbf{Correct Answer: C}

    \item A baseball player got two hits for three times at bat. This could be represented by 2/3. Which decimal most closely represents this?
    \begin{itemize}[itemsep=0.01pt]
        \item[A.] 0.20
        \item[B.] 0.33
        \item[C.] 0.50
        \item[D.] 0.67
        \item[E.] 0.80
    \end{itemize}
    \textbf{Correct Answer: D}

    \item If P = M + N, which of the following will be true? (a) N = P - M  (b) P - N = M  (c) N + M = P.
    \begin{itemize}[itemsep=0.01pt]
        \item[A.] Only (a) is true
        \item[B.] (a) and (b) only
        \item[C.] (a) and (c) only
        \item[D.] (b) and (c) only
        \item[E.] (a), (b), and (c) are all true
    \end{itemize}
    \textbf{Correct Answer: E}

    \item The hands of a clock form an obtuse angle at \_\_\_\_ o'clock.
    \begin{itemize}[itemsep=0.01pt]
        \item[A.] 1 o'clock
        \item[B.] 2 o'clock
        \item[C.] 3 o'clock
        \item[D.] 4 o'clock
        \item[E.] 6 o'clock
    \end{itemize}
    \textbf{Correct Answer: D}

    \item Bridget buys a packet containing 9-cent and 13-cent stamps for \$2.65. If there are 25 stamps in the packet, how many are 13-cent stamps?
    \begin{itemize}[itemsep=0.01pt]
        \item[A.] 5
        \item[B.] 8
        \item[C.] 10
        \item[D.] 12
        \item[E.] 15
    \end{itemize}
    \textbf{Correct Answer: C}

    \item On a certain map, 7/8 inch represents 200 miles. How far apart are two towns whose distance apart on the map is 3 half inches?
    \begin{itemize}[itemsep=0.01pt]
        \item[A.] about 217 miles
        \item[B.] about 427 miles
        \item[C.] about 395 miles
        \item[D.] about 343 miles
        \item[E.] about 1,000 miles
    \end{itemize}
    \textbf{Correct Answer: D}

    \item Fred's bill for some household supplies was \$13.64. If he paid for the items with a \$20 bill how much change should he receive?
    \begin{itemize}[itemsep=0.01pt]
        \item[A.] \$5.36
        \item[B.] \$6.14
        \item[C.] \$6.36
        \item[D.] \$6.46
        \item[E.] \$7.36
    \end{itemize}
    \textbf{Correct Answer: C}

    \item Some people suggest that the following formula be used to determine the average weight for boys between the ages of 1 and 7: W = 17 + 5A, where W is weight in pounds and A is age in years. According to this formula, for each year older a boy gets, should his weight become more or less and by how much?
    \begin{itemize}[itemsep=0.01pt]
        \item[A.] Weight decreases by 5 pounds each year
        \item[B.] Weight decreases by 17 pounds each year
        \item[C.] Weight stays the same each year
        \item[D.] Weight increases by 5 pounds each year
        \item[E.] Weight increases by 17 pounds each year
    \end{itemize}
    \textbf{Correct Answer: D}

    \item Five spelling tests are to be given to Mary's class. Each test has a value of 25 points. Mary's average for the first four tests is 15. What is the highest possible average she can have on all five tests?
    \begin{itemize}[itemsep=0.01pt]
        \item[A.] 15
        \item[B.] 16
        \item[C.] 17
        \item[D.] 18
        \item[E.] 20
    \end{itemize}
    \textbf{Correct Answer: C}

    \item Compute: 3 4/5 - 1/2.
    \begin{itemize}[itemsep=0.01pt]
        \item[A.] 3 1/2
        \item[B.] 3 3/10
        \item[C.] 3 2/5
        \item[D.] 2 3/10
        \item[E.] 2 4/5
    \end{itemize}
    \textbf{Correct Answer: B}

    \item In an auditorium, the chairs are usually arranged so that there are x rows and y seats in a row. For a popular speaker, an extra row is added and an extra seat is added to every row, so there are x + 1 rows and y + 1 seats per row. Multiply (x + 1)(y + 1).
    \begin{itemize}[itemsep=0.01pt]
        \item[A.] xy + 1
        \item[B.] xy + x + 1
        \item[C.] xy + x + y + 1
        \item[D.] xy + y + 1
        \item[E.] xy + 2
    \end{itemize}
    \textbf{Correct Answer: C}

    \item A ferris wheel measures 80 feet in circumference. The distance on the circle between two of the seats is 10 feet. Find the measure in degrees of the central angle whose rays support the two seats.
    \begin{itemize}[itemsep=0.01pt]
        \item[A.] 30$^\circ$
        \item[B.] 36$^\circ$
        \item[C.] 40$^\circ$
        \item[D.] 45$^\circ$
        \item[E.] 60$^\circ$
    \end{itemize}
    \textbf{Correct Answer: D}

    \item Set up the problem needed to find the number in the expression 'six less than twice 4 5/6'.
    \begin{itemize}[itemsep=0.01pt]
        \item[A.] 6 - 2(4 5/6)
        \item[B.] 2(4 5/6) - 6
        \item[C.] 4 5/6 - 2$\cdot$6
        \item[D.] 2(6 - 4 5/6)
        \item[E.] 2(4 5/6 + 6)
    \end{itemize}
    \textbf{Correct Answer: B}

    \item Two triangles are similar. The corresponding sides are proportional and AC / BD = XZ / YZ. If AC = 1.7, BC = 2, and XZ = 5.1, find YZ.
    \begin{itemize}[itemsep=0.01pt]
        \item[A.] 2
        \item[B.] 3
        \item[C.] 4
        \item[D.] 5
        \item[E.] 6
    \end{itemize}
    \textbf{Correct Answer: E}
\end{enumerate}

\begin{lstlisting}[language=json, caption={Task 4 specific prompt instructions for mathematical problem-solving, enforcing the selection of a single multiple-choice option, step-by-step reasoning, and a self-reported confidence score.}, label={lst:task4_prompt}] """ You are an assistant generating structured JSON data for research.
Your response MUST be a single valid JSON object with the structure shown below.

ROLE:
{role_instructions}

TASK - Mathematics problem solving (MSESR MCQs).

For EACH problem:
- Pick exactly one 'chosen_option' from ["A","B","C","D","E"].
- Provide a short 'reasoning' (max ~40 words) explaining how the answer was obtained.
- Provide a numerical estimation of how confident you felt in your answer from 1 (not confident) to 5 (very confident).

OUTPUT FORMAT (EXAMPLE ONLY - replace placeholders):

{schema_json}

PROBLEMS AND OPTIONS:
{mcq_block}

Return ONLY the JSON object, no extra commentary. """

\end{lstlisting}









\section{Hardware infrastructure and implemented computational resources}\label{app:B}

\begin{table}[h!]
\caption{Hardware infrastructure and implemented computational resources. Data generation and feature extraction were distributed across six dedicated machines. Large Language Model (LLM) inference required GPU-enabled systems, whereas TFMN and BFMN construction and validation relied on CPU-intensive architectures. Runtime values indicate total processing time per model or run.}
\label{tab:hardware}
\centering
\renewcommand{\arraystretch}{1.3}

\resizebox{\textwidth}{!}{
\begin{tabular}{|p{3cm}|p{3cm}|p{7cm}|p{3cm}|}
\toprule
\rowcolor[HTML]{E67E22}
\textcolor{white}{\textbf{Process}} &
\textcolor{white}{\textbf{Machine}} &
\textcolor{white}{\textbf{Machine Specifications}} &
\textcolor{white}{\textbf{Total Runtime (min.)}} \\
\midrule

\rowcolor[HTML]{FEF0E7}
\multirow{2}{*}{LLM Data Generation}
& HP Z1 G9 Tower  
& GPU: GTX 3060 (12 GB RSX); CPU: Intel Core i7-13700, 16 cores; RAM: 32 GB;
& \multirow{3}{*}{ $\sim$25,320 (422 hours)} \\
\cline{2-3}

\rowcolor[HTML]{FEF0E7}
& {TUF ASUS Tower}  
& GPU: GTX 3060 (12 GB RSX); CPU: Intel Core i9-11900K, 8 cores; RAM: 32 GB;
& \\
\cline{2-3}

\rowcolor[HTML]{FEF0E7}
& Dell Alienware  Aurora R16 
& GPU: NVIDIA RTX 4090 (24 GB GDDR6X); CPU: Intel Core i7-14700F, 20 cores; RAM: 32 GB DDR5; Storage: 2 TB NVMe SSD 

& \\
\cline{2-3}

\rowcolor[HTML]{FEF0E7}
& Dell PowerEdge  R7525 
& GPU: NVIDIA L40 (48 GB); CPU: 2$\times$ AMD EPYC 7313, 32 cores total; RAM: 128 GB; Storage: 2$\times$960 GB SATA SSD 
& \\

\midrule

\multirow{2}{*}{Data Cleaning}
& MacBook Pro (2024)
& CPU: Apple Silicon M4 Pro chip (12 cores); RAM: 24 GB; Storage: 512 GB SSD; GPU: Apple-designed integrated graphics (16 cores); MacOS 26 Tahoe
& \multirow{2}{*}{$\sim$300}\\
\midrule

\rowcolor[HTML]{FEF0E7}
\multirow{2}{*}{\parbox{2.8cm}{Data Preprocessing \&}}
& Galaxy book 4 
& CPU: 13Gen Intel core i5 1335U; RAM: 16 GB; Storage: 1TB SSD; GPU: iRIS(R) Xe Graphics 
& $\sim$2100 \\
\cline{2-3}

\rowcolor[HTML]{FEF0E7}
\multirow{2}{*}{\parbox{2.8cm}{TFMN Network Analysis \& Validation \& Results}}
& M75s-1 Desktop (ThinkCentre)
& CPU: MD Ryzen 5 PRO 3400G(4 cores / 8 threads); RAM: 16 GB; Storage: SSD(NVMe);  GPU: AMD Radeon Vega 11; Windows 11 
&\\
\cline{2-3}

\rowcolor[HTML]{FEF0E7}
& MacBook Air (2020)
& CPU: Apple Silicon M1 chip (8 cores); RAM: 8 GB; Storage: 256 GB;
& \\
\cline{2-3}

\cline{2-3}
\rowcolor[HTML]{FEF0E7}
& DELL Vostro 3590
& CPU: Intel Core i5-10210U (4 cores); RAM: 8 GB; Storage: 256 GB NVMe SSD; GPU: Intel UHD Graphics; Windows 11 64-bit
& \\

\cline{2-3}
\rowcolor[HTML]{FEF0E7}
& MacBook Pro (2024)
& CPU: Apple Silicon M4 Pro chip (12 cores); RAM: 24 GB; Storage: 512 GB SSD; GPU: Apple-designed integrated graphics (16 cores); MacOS 26 Tahoe
& \\
\bottomrule

\end{tabular}
}

\vspace{0.6em}
\footnotesize
\end{table}

\section{Data pre-processing pipeline}\label{app:C}

\begin{figure}[h!]
    \centering
    \includegraphics[width=1\linewidth]{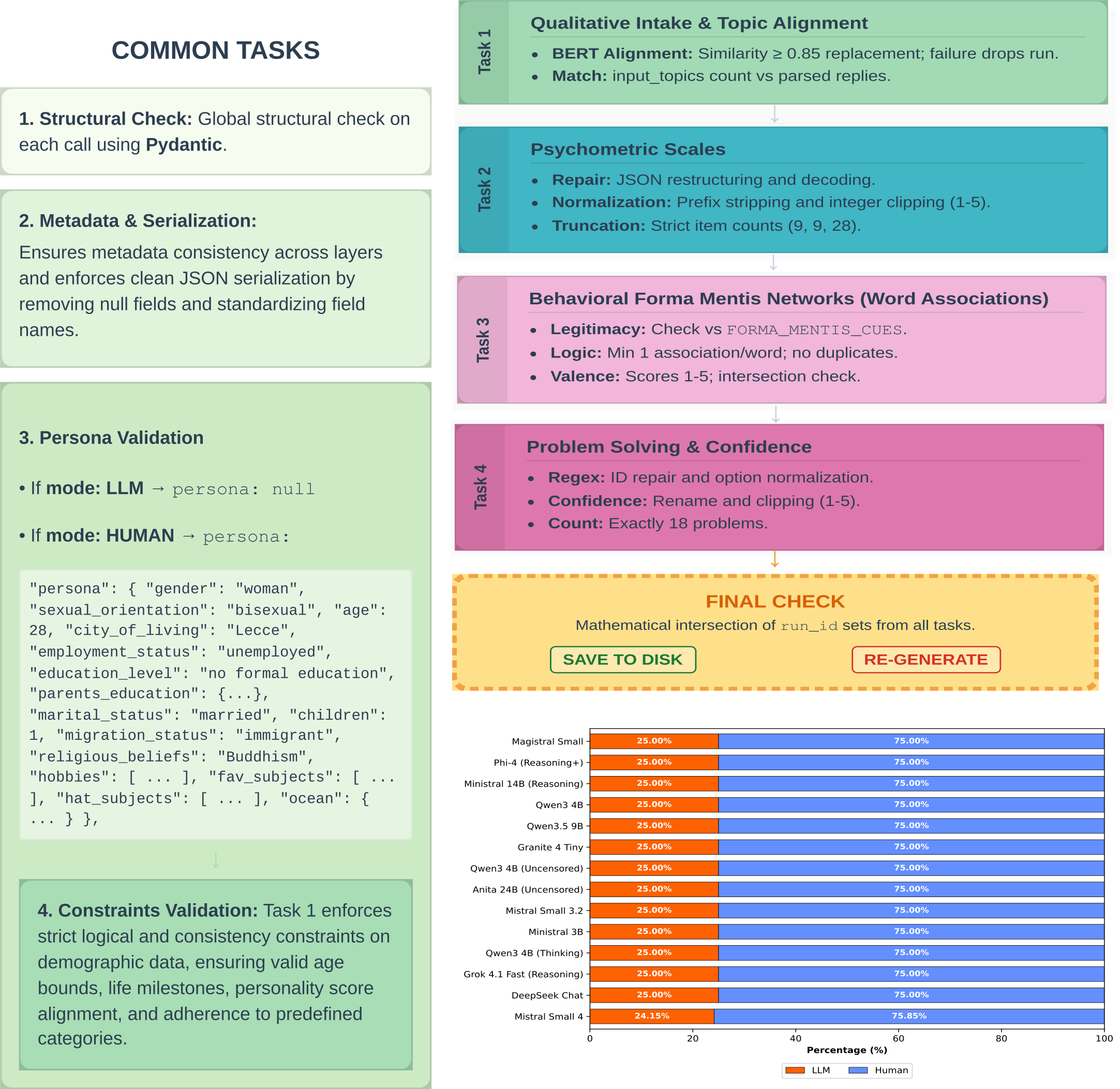}
    \caption{Overview of the data pre-processing pipeline. The pipeline applies a set of common checks to all runs (structural analysis, metadata serialization, persona consistency, and demographic constraints), followed by four task-specific analysis targeting qualitative responses (Task~1), psychometric scale ratings (Task~2), semantic word associations (Task~3), and mathematical problem-solving outputs (Task~4). A final cross-task integrity check precedes data serialization. The final plot shows the proportion of personas that have been simulated with the LLM and and human-shadow modes for all LLMs considered in the study.}
    \label{fig:data_preprocessing}
\end{figure}

\section{Supplementary Data Records}\label{app:D}

This appendix documents the structure of the dataset generated 
through the experimental pipeline. The two tables below provide 
a comprehensive reference for all variables recorded across the 
four tasks.

Table~\ref{tab:common_variables} describes the variables shared 
across all runs, including task identifiers, model metadata, and, when the LLM was prompted to generate a human shadow, the full set of socio-demographic and psychological attributes used to instantiate each human shadow. While the base sampling weights for these attributes (such as age, gender, sexual orientation, and others) are summarized directly within the table, the full generative code, detailing all the status-consistent constraints and complex conditional logic, is available and can be inspected in the project repository hosted on \href{https://github.com/MassimoStel/MEDS}{GitHub}.

Table~\ref{tab:llm_perspectives} details the task-specific output variables, documenting the structured fields produced by the model for each of the four calls: the interview responses (Task~1), the psychometric scale ratings (Task~2), the semantic association maps (Task~3), and the mathematical problem-solving scores (Task~4).

\begin{table}[H]
\caption{Common variables recorded across all experimental runs, 
including task metadata, model information, and the socio-demographic 
and psychological attributes used to define human-shadows, along 
with their base sampling weights. The generative code containing the full conditional logic and sampling details is openly available on \href{https://github.com/MassimoStel/MEDS}{GitHub}.}
\label{tab:common_variables}
\centering
\footnotesize
\renewcommand{\arraystretch}{1.3}
\begin{tabularx}{\textwidth}{|l|l|X|}
\hline
\rowcolor[HTML]{E67E22}
\textcolor{white}{\textbf{Variable}} & 
\textcolor{white}{\textbf{Type}} & 
\textcolor{white}{\textbf{Description}} \\ \hline

\multicolumn{3}{|l|}{\cellcolor[HTML]{F2F2F2}\textbf{Common variables}} \\ \hline

\rowcolor[HTML]{FEF0E7}
\texttt{run\_id} & string & Unique identifier for each experimental run. \\ \hline

\texttt{call\_name} & string & Label identifying the specific pipeline call: 
\texttt{call1\_topics}, \texttt{call2\_scales}, 
\texttt{call3\_forma\_mentis\_batch1}, 
\texttt{call3\_forma\_mentis\_batch2}, 
\texttt{call4\_msesr\_mcq}. \\ \hline

\rowcolor[HTML]{FEF0E7}
\texttt{task} & string & Identifier for the experimental task: 
\texttt{call1}, \texttt{call2}, \texttt{call3}, \texttt{call4}. \\ \hline

\texttt{mode} & string & Experimental condition:  \texttt{human} 
(persona simulation), \texttt{LLM} (baseline LLM response). \\ \hline

\rowcolor[HTML]{FEF0E7}
\texttt{model} & string & Name of the LLM used to generate the record. \\ \hline

\texttt{base\_url} & string & API endpoint used to query the LLM. \\ \hline

\rowcolor[HTML]{FEF0E7}
\texttt{persona} & dict / null & When \texttt{mode} is \texttt{human}: a 
dictionary of socio-demographic and psychological attributes (see below). 
When \texttt{mode} is \texttt{LLM}: \texttt{null}. \\ \hline

\multicolumn{3}{|l|}{\cellcolor[HTML]{F2F2F2}\textbf{Socio-demographic 
and psychological variables (persona)}} \\ \hline

\texttt{age} & integer & Age of the simulated persona. Weighted sampling: 
18, 19, 20 (9.1\% each); 21--25, 30, 35, 40, 50 (6.1\% each); 
26--29, 60, 70 (3.0\% each). \\ \hline

\rowcolor[HTML]{FEF0E7}
\texttt{gender} & string & \texttt{man} (38.0\%), \texttt{woman} (38.0\%), \texttt{transgender} (12.5\%), \texttt{genderqueer} (4.5\%), \texttt{agender} (4.5\%), \texttt{non-binary} (2.5\%). \\ \hline

\texttt{sexual\_orientation} & string & \texttt{heterosexual} (62.5\%),
\texttt{homosexual} (12.5\%), \texttt{bisexual} (12.5\%), \texttt{asexual} (12.5\%). \\ \hline

\rowcolor[HTML]{FEF0E7}
\texttt{city\_of\_living} & string & City of residence uniformly distributed (8.3\% each). Italian cities:
Rome, Milan, Naples, Bari, Lecce, Bologna. US cities: New York,
Los Angeles, Kansas City, Philadelphia, Chicago, Washington~DC. \\ \hline

\texttt{employment\_status} & string & Conditionally sampled based on both age and education level. Student status is heavily weighted for younger profiles, professional employment combinations vary by degree level, and retirement is strictly constrained to older personas. Complete list: \texttt{self-employed}, 
\texttt{employed full-time}, \texttt{employed part-time}, 
\texttt{full-time student}, \texttt{part-time student}, \texttt{unemployed}, \texttt{retired}. \\ \hline

\rowcolor[HTML]{FEF0E7}
\texttt{education\_level} & string & Conditionally sampled based on age. Younger profiles ($\le$ 21) are restricted to secondary, vocational, or bachelor's levels, while advanced degrees become available and weighted differently for older profiles.
Complete list: \texttt{no formal education}, 
\texttt{primary school}, \texttt{lower secondary}, 
\texttt{upper secondary}, \texttt{vocational diploma}, 
\texttt{bachelor's degree}, \texttt{master's degree}, \texttt{PhD}. \\ \hline

\texttt{parents\_education} & dict & Dictionary with keys \texttt{parent\_1} 
and \texttt{parent\_2}, each mapped to an \texttt{education\_level} 
value as defined above. \\ \hline

\rowcolor[HTML]{FEF0E7}
\texttt{marital\_status} & string & Conditionally sampled based on age. Younger profiles ($<$ 25) are heavily weighted toward being \texttt{single} or \texttt{in a relationship}, while older profiles are progressively weighted toward \texttt{married}, \texttt{divorced}, or \texttt{widowed}. \\ \hline

\texttt{children} & integer & Conditionally sampled (0, 1, or 2) based on age, marital status, and sexual orientation (e.g., strong base weight for 0 children if single and $<$22 yrs old; weights dynamically penalized for asexual profiles). \\ \hline

\rowcolor[HTML]{FEF0E7}
\texttt{migration\_status} & string & Uniformly distributed (33.3\% each): \texttt{native-born Italian}, 
\texttt{native-born American}, \texttt{immigrant}. \\ \hline

\texttt{religious\_beliefs} & string &  \texttt{Christianity} (26.7\%),
\texttt{Islam} (26.7\%), \texttt{Hinduism} (13.3\%), \texttt{Atheist} (13.3\%), \texttt{Buddhism} (6.7\%), \texttt{Judaism} (6.7\%), \texttt{Agnostic} (6.7\%). \\ \hline

\rowcolor[HTML]{FEF0E7}
\texttt{hobbies} & list[string] & Four hobbies sampled from: hiking, 
reading novels, baking, football, running, yoga, board games, 
photography, gardening, volunteering, cinema, classical music, painting, 
cycling, videogames, travel, cooking, birdwatching, knitting, 
rock climbing. \\ \hline

\texttt{fav\_subjects} & list[string] & Two favourite subjects sampled 
from: maths, physics, chemistry, biology, computer science, philosophy, 
history, art, economics, literature. \\ \hline

\rowcolor[HTML]{FEF0E7}
\texttt{hated\_subjects} & list[string] & Two disliked subjects sampled 
from the same list as \texttt{fav\_subjects}. \\ \hline

\texttt{ocean} & dict & Big Five personality profile. Each of the five 
dimensions (\texttt{openness}, \texttt{conscientiousness}, 
\texttt{extraversion}, \texttt{agreeableness}, \texttt{neuroticism}) 
is represented by two keys: \texttt{score} (integer in $[0, 100]$) and 
\texttt{level} (\texttt{low} if score $<33$; \texttt{moderate} if 
$33 \le$ score $\le 66$; \texttt{high} if score $> 66$). \\ \hline

\end{tabularx}
\end{table}

\begin{table}[H]
\caption{Task-specific output variables produced by the model 
for each of the four experimental tasks.}
\label{tab:llm_perspectives}
\centering
\footnotesize
\renewcommand{\arraystretch}{1.3}
\begin{tabularx}{\textwidth}{|l|l|X|}
\hline
\rowcolor[HTML]{E67E22}
\textcolor{white}{\textbf{Variable}} & 
\textcolor{white}{\textbf{Type}} & 
\textcolor{white}{\textbf{Description}} \\ \hline

\multicolumn{3}{|l|}{\cellcolor[HTML]{F2F2F2}\textbf{Task 1 — Interview}} \\ \hline

\rowcolor[HTML]{FEF0E7}
\texttt{input\_topics} & list[string] & 
The seven interview questions posed to the model. \\ \hline

\texttt{raw\_output} & string & 
Unparsed response to all questions, concatenated as a single string. \\ \hline

\rowcolor[HTML]{FEF0E7}
\texttt{parsed} & dict & 
Structured output containing \texttt{mode} and \texttt{replies}. \\ \hline

\texttt{replies} & dict & 
Each question acts as a key; the associated value is the model's answer. \\ \hline

\rowcolor[HTML]{FEF0E7}
\texttt{reasoning\_summary} & string & 
Model's explanation of why it provided those specific answers. \\ \hline

\multicolumn{3}{|l|}{\cellcolor[HTML]{F2F2F2}\textbf{Task 2 — Psychometric Scales}} \\ \hline

\texttt{raw\_outputs} & dict & 
Raw JSON responses for each of the three scales 
(\texttt{MSES}, \texttt{amas}, \texttt{mseaq}). \\ \hline

\rowcolor[HTML]{FEF0E7}
\texttt{parsed} & dict & 
Structured output containing \texttt{mode}, \texttt{persona}, 
and \texttt{scales}. \\ \hline

\texttt{scales} & dict & 
Three sub-objects, one per scale, each containing \texttt{items}. \\ \hline

\rowcolor[HTML]{FEF0E7}
\texttt{items} & dict & 
Individual scale items. Each item contains \texttt{rating} 
and \texttt{why}. \\ \hline

\texttt{rating} & integer & 
Quantitative score assigned to the item, in the range $[1, 5]$. \\ \hline

\rowcolor[HTML]{FEF0E7}
\texttt{why} & string & 
Qualitative reasoning provided by the model for the chosen rating. \\ \hline

\multicolumn{3}{|l|}{\cellcolor[HTML]{F2F2F2}\textbf{Task 3 — Semantic Associations}} \\ \hline

\texttt{cue\_words} & list[string] & 
The 50 target words used as primary stimuli for semantic association, see \ref{app:A_task3} \\ \hline

\rowcolor[HTML]{FEF0E7}
\texttt{parsed} & dict & 
Root container split into \texttt{forma\_mentis} 
and a persona verification object. \\ \hline

\texttt{forma\_mentis} & dict & 
One entry per cue word, each containing 
\texttt{associations} and \texttt{valence}. \\ \hline

\rowcolor[HTML]{FEF0E7}
\texttt{associations} & list[string] & 
Up to three words the model associates with the cue word. \\ \hline

\texttt{valence} & integer & 
Sentiment score for the cue word and its associations, 
in the range $[1, 5]$ (1--2: negative; 3: neutral; 4--5: positive). \\ \hline

\multicolumn{3}{|l|}{\cellcolor[HTML]{F2F2F2}\textbf{Task 4 — Mathematical Problem Solving}} \\ \hline

\rowcolor[HTML]{FEF0E7}
\texttt{raw\_output} & string & 
Full unparsed response from the model. \\ \hline

\texttt{parsed} & dict & 
Structured output containing \texttt{mode} 
and \texttt{msesr\_problem\_solving}. \\ \hline

\rowcolor[HTML]{FEF0E7}
\texttt{msesr\_problem\_solving} & dict & 
Results keyed by question number (1--18), each containing 
\texttt{chosen\_option}, \texttt{reasoning}, 
and \texttt{confidence\_score}. \\ \hline

\texttt{chosen\_option} & string & 
Selected multiple-choice answer (e.g., \texttt{A}, \texttt{B}). \\ \hline

\rowcolor[HTML]{FEF0E7}
\texttt{reasoning} & string & 
Step-by-step mathematical justification for the chosen answer. \\ \hline

\texttt{confidence\_score} & integer & 
Model's self-reported certainty, in the range $[1, 5]$. \\ \hline

\texttt{reasoning\_summary} & string & 
Global overview of the model's answers and any corrections made. \\ \hline

\end{tabularx}
\end{table}



\bibliography{main}